\documentclass[lettersize,journal]{IEEEtran}
\usepackage{amsmath,amsfonts}
\usepackage{algorithm}
\usepackage{algpseudocode}
\usepackage{amsmath}
\usepackage{hyperref}
\usepackage{array}
\usepackage[caption=false,font=normalsize,labelfont=sf,textfont=sf]{subfig}
\usepackage{textcomp}
\usepackage{stfloats}
\usepackage{url}
\usepackage{verbatim}
\usepackage{graphicx}
\usepackage{cite}
\usepackage{float}

\usepackage{enumitem}

\begin{document}

\title{Biologically Inspired Swarm Dynamic Target Tracking and Obstacle Avoidance}

\author{Lucas Page \\ The University of New South Wales Canberra at the Australian Defence Force Academy, Campbell, ACT, 
2612, Australia }

\maketitle

\begin{abstract}
This study proposes a novel artificial intelligence (AI) driven flight computer, integrating an online free-retraining-prediction model, a swarm control, and an obstacle avoidance strategy, to track dynamic targets using a distributed drone swarm for military applications. To enable dynamic target tracking the swarm requires a trajectory prediction capability to achieve intercept allowing for the tracking of rapid maneuvers and movements while maintaining efficient path planning. Traditional predicative methods such as curve fitting or Long Short Term Memory (LSTM) have low robustness and struggle with dynamic target tracking in the short term due to slow convergence of single agent-based trajectory prediction and often require extensive offline training or tuning to be effective. Consequently, this paper introduces a novel robust adaptive bidirectional fuzzy brain emotional learning prediction (BFBEL-P) methodology to address these challenges. The controller integrates a fuzzy interface, a neural network enabling rapid adaption, predictive capability and multi-agent solving enabling multiple solutions to be aggregated to achieve rapid convergence times and high accuracy in both the short and long term. This was verified through the use of numerical simulations seeing complex trajectory being predicted and tracked by a swarm of drones. These simulations show improved adaptability and accuracy to state of the art methods in the short term and strong results over long time domains, enabling accurate swarm target tracking and predictive capability.
\end{abstract}

\section*{Nomenclature}

\begin{description}[labelwidth=1.5cm, labelsep=1cm, leftmargin=4cm]
  \item[$K_p$] Proportional Constant
  \item[$K_i$] Integral Constant
  \item[$K_d$] Derivative Constant
  \item[$R_C$] Cohesion Rate 
  \item[$R_S$] Separation Rate 
  \item[$R_A$] Alignment Rate 
  \item[$e(t)$] Error as a function of time
  \item[$u(t)$] Control input as a function of time
  \item[$x, y, z$] Drone Coordinates
  \item[$x_o, y_o, z_o$] Other Drone Coordinates
  \item[$x_t, y_t, z_t$] Target Coordinates
  \item[$x_s, y_s, z_s$] Swarm Body Coordinates
  \item[$p_{ij}$] Fuzzy Membership Function
  \item[$v_{ij}$] Amygdala Weight Function
  \item[$w_{ij}$] Orbitofrontal Cortex Weight Function
  \item[$\bar{x}$] Rolling Average
  \item[$s$] Transfer Function
  \item[$a, o$] Amygdala and Orbitofrontal Cortex terms
  \item[$R$] Reward Function
  \item[$q$] Gain Vector
  \item[$u_{BFBEL}$] BFBEL-P Response
  \item[$\alpha$, $\beta$] Learning Rates
  \item[$\textbf{i}$] Input Series
  \item[$\textbf{c}$] Comparison Series
  \item[$\textbf{f}$] Prediction Series
  \item[$\sigma$] Variance
  \item[$\varsigma$] Mean
  \item[$\eta$] Obstacle Repulsive Strength Scaling Factor
  \item[$\rho_0$] Obstacle Influence Distance
\end{description}

\section{Introduction}
\IEEEPARstart{T}{he} onset of War in Ukraine has seen a dramatic shift in combined arms warfare through the large scale introduction of the drones to the modern battlefield \cite{ISOW_Restore_manuver}. These systems are pioneering new techniques in reconnaissance, close air support and strategic strike capabilities \cite{Low_Cost_Mass_Produced_Drones_a_Tactical_Advantage_Citation}, however are naturally limited to one pilot controlling one drone. Current civilian drone swarms utilize formation control, where each drone in the swarm occupies a predetermined location and remains there for the duration of the activity. Such a control measure is perfect for a single use case such as a light show, however it is extremely limited for use in a dynamic contested environments. This is due to the challenges of calculating fixed relative position drone motion with complex maneuvering. \\

In contrast, a distributed drone swarm can be used where drones are not fixed in location but rather remain distributed in a region, swarmed around a centralized point which can be driven to locations. Compared to single drones a swarm has incredibly high robustness due to it being able to complete the mission even in the event of drone losses to hardware or combat interference, furthermore they can be used for a multitude of simultaneous task through having several specialist drones working in unison \cite{Multi-gas-source-localization-by-flocking-robots}. Furthermore, the rapid and accelerating rise of autonomous systems driven by enhanced Artificial Intelligence (AI) models can act as an enabler, expanding the scope and assisting in decision making seeing a raise in the capability of assets \cite{rickli2023artificial}. Consequently, this thesis focuses on the development of a distributed drone swarm flight computer enabling dynamic target tracking and obstacle avoidance through the use of novel AI target prediction models to achieve online tracking and strike adaptability in dynamic environments, as a method to control a swarm of UAV assets for their implementation on the modern battlefield.

\section{Aims and Significance}
To date, a large-scale distributed controlled drone swarm has not been utilized on warlike operations, however the war in Ukraine has praised drones for their cheap cost, versatility, and minimal training requirements, allowing both conventional and unconventional forces to project significant firepower from concealed locations \cite{Swarms_of_Trouble_The_Hidden_Threat_of_Consumer_UAVs__Citation} without risking a human operator. Additionally with an extremely low opportunity cost compared to conventional guided munitions the barrier for entry has dramatically dropped for such a system rapidly diminishing the capability gap boasted by modern conventional military's \cite{militant_drone_use}, reinforced by Former Chief of the Royal Australian Air Force, Air Marshal Chipman announcements that Australia needs to invest in low-cost mass produced drones \cite{Low_Cost_Mass_Produced_Drones_a_Tactical_Advantage_Citation} to remain combat effective in the future. Furthermore the center for strategic and international studies identified a critical shortage of modern United States (US) based missile systems that are currently ready for use in a full scale war in the Taiwan straight \cite{no_missile_us}, indicating that there would be critical shortages within a week. This supply shortage and lack of surge capacity affects not just the US but all coalition nations and drones have the potential amongst other weapon systems to help bridge that capability gap \cite{us_army_drone_gap}.\\

\begin{figure}[H]
\includegraphics[width=3.6in]{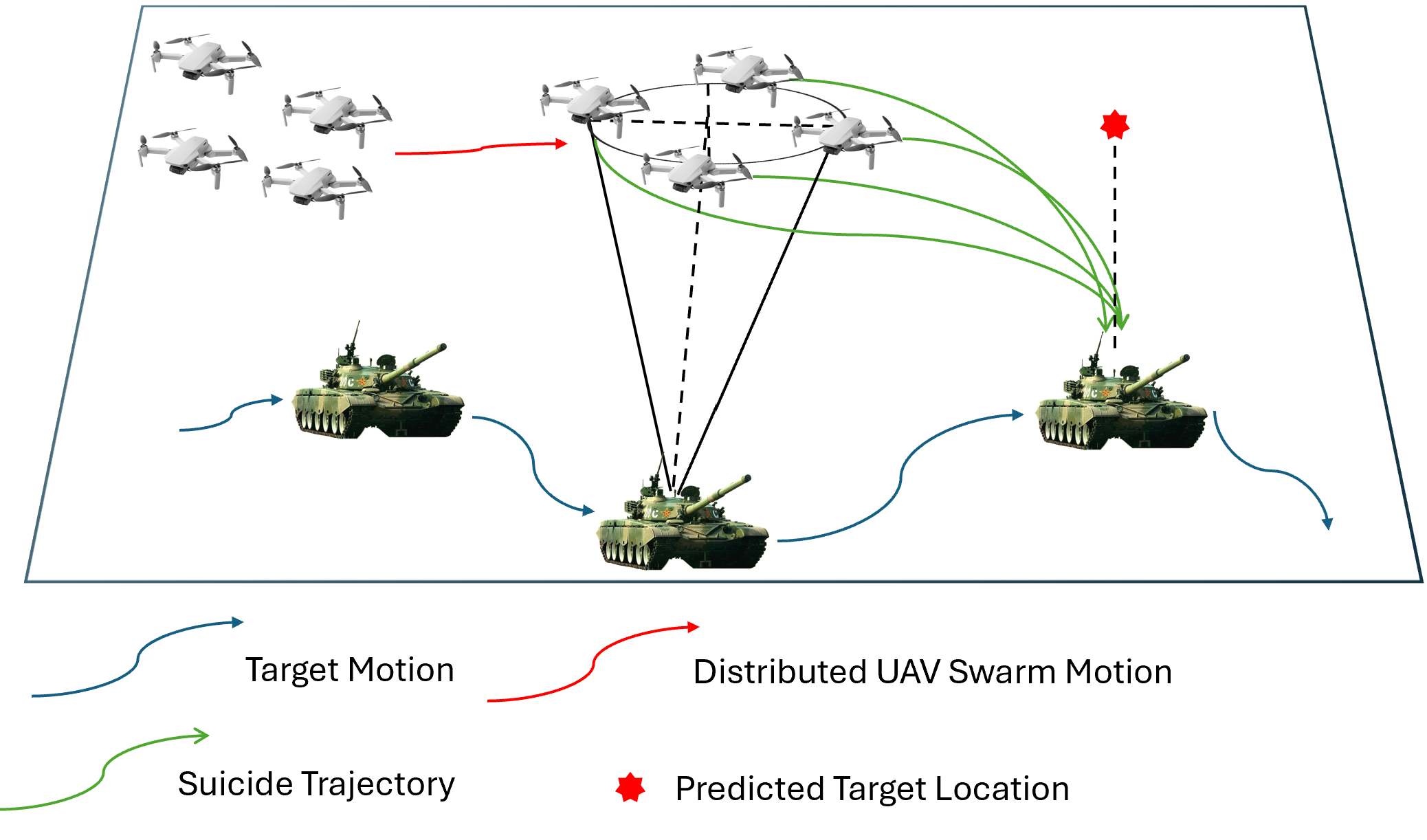}
\caption{Possible Drone Swarm Attack Diagram.}
\label{fig:drone_attack}
\end{figure}

Consequently, as the use of drones becomes more relevant in warfare, they can be used as a loitering munition to rapidly strike targets \cite{ADF_Shallow_Loitering}; this would see a swarm of drones controlled by a single operator, providing direction join fires on discrete targets such as in \autoref{fig:drone_attack}. Furthermore a swarm of drones has the ability to have a multiple roles fulfilled by the one swarm with various drones taking different roles. The significance of this versatility is that drone swarms with dynamic target tracking and obstacle avoidance have the potential to act as a major avenue for the conduct of warfare \cite{dowd2013drone} in the future and is a critical venture to be understood, implemented and appreciated. Therefore, the aims of this study is to realize this capability in the form of a scalable swarm that can dynamically predict and track a targets motion despite not knowing any prior information about what the target is or the drones surroundings. To accomplish this an AI-driven flight computer will be developed to effectively coordinate and control the drones, ensuring efficient tracking of dynamic targets while navigating around obstacles. The proposed approach involves simulating drones within a Robotic Operating System (ROS) Gazebo simulation environment, chosen for its ability to provide realistic physics, robust data collection capabilities and seamless transition into real-world drone operations.

\section{Related Works}
This thesis is built upon past research and supporting literature, as such a review is conducted on the selection of Single vs Multi Drone systems \autoref{lit:Single_vs_multi_drone}, formation vs swarm control \autoref{lit:formation_vs_swarm}, obstacle avoidance measures \autoref{lit:obstacle_avoid}, predictive methodologies \autoref{lit:predictive_models}, and data fusion processes \autoref{lit:data_fusion}. The review justifies the decision to use multi drone distributed control structures with APF driven obstacle avoidance driven by novel Artificial Intelligence trajectory prediction encompassing rolling average data fusion approaches. This sees the swarm utilize state-of-the-art approaches and novel architecture for the development of swarm capable for use in a combat zone.

\subsection{Single vs Multi Drone Swarms}
\label{lit:Single_vs_multi_drone}
A singular UAV system is naturally limited by the fact it is a single drone operating and requires a human operator for each UAV agent subsequently limiting their coverage, capabilities and critical robustness of the system. In a contested and dynamic environment, there is a high possibility of a node (drone) loss, such an outcome would inhibit the completion of the mission or task. Furthermore, a single drone only has the capability to complete one primary task, a swarm can have multiple specialist drones working together, acting as a force multiplier \cite{drone_force_multiplyer} to improve overall swarm effectiveness. Chen \cite{chen2020toward} demonstrates that swarms can collaboratively complete tasks with higher efficiency and robustness. Critical to this however is the networking technologies so that the swarm can achieve consensus, collaborate and information share. A swarm of drones has the capability to remain in the air loitering waiting for a mission and then execute the mission autonomously with sufficient numbers for combat effectiveness without having to scramble additional UAV assets. This allowing for a significantly quicker response to a dynamic environment. Furthermore, the US Army War College \cite{dowd2013drone} has shown a 1,200 percent increase in UAV combat air patrols since 2005 and "The Air Force envisions deploying swarms of drones networked together to operate in a variety of lethal and non-lethal missions at the command of a single pilot”\cite{dowd2013drone}. Subsequently, the future usage of drone swarms for large-scale combat operations is inevitable, and consequently, this thesis and review will focus on the implementation of multi-agent drone swarms.

\vfill
\subsection{Formation vs Swarm Control}
\label{lit:formation_vs_swarm}
\begin{figure}[H]
\begin{centering}
\includegraphics[width=3.4in]{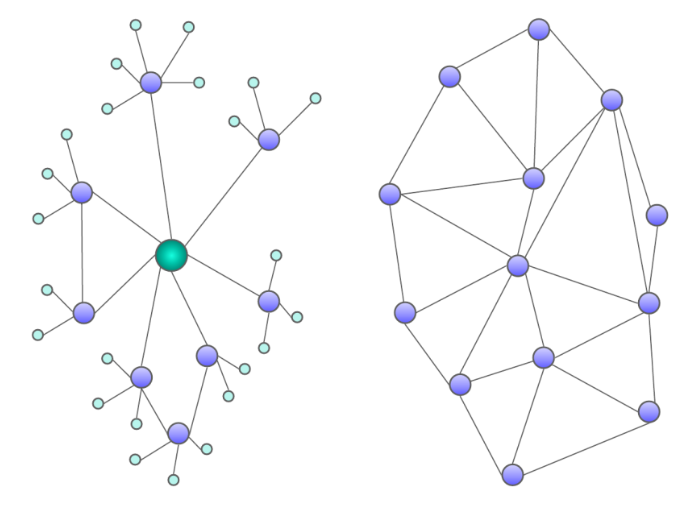}
\caption{Formation (left) vs Swarm (right) Architecture.\cite{s23218766}}
\label{fig:SwarmVScentral}
\end{centering}
\end{figure}

 A coalition of drones requires the ability to achieve consensus on its position in order to dictate swarm decision making. Traditionally formation control was used to keep the drones in a fixed position relative to each other to complete its objectives. However such a system has high computational complexity for maneuvers and very low robustness with node loss \cite{CHEN2021504}. More modern research is focused on the use of distributed network for swarm control. This sees the drones operate as a collective entity and provides significantly improved robustness and improved performance. A comparison of these systems is displayed in \autoref{fig:SwarmVScentral}.

 Formation control can be achieved through control schemes such as leader follower, where a local leader controls a sub group of drones where all leaders are connected nodes. Critical to this process is achieving robustness for node failure and control in large scale drone formations. Node failure can be partially mitigated through using average consensus and grey prediction for leaderless drones \cite{chen2020achieving}. Such systems have demonstrated a higher degree of robustness and lower computational complexity with no significant changes to convergence rates with large formations. Alternatively a large scale drone formation framework can be achieved via global coordination through local interaction combined with a consensus algorithm considering merging behaviour to achieve leader follower consensus in a short time \cite{chen2024fast}. Such a system sees drones interact with adjacent drones to achieve the coordination, these algorithms are then discredited via the Runge-Kutta method allowing the algorithms to be effective in large scale swarms\cite{chen2024fast} without substantial computational strain. Despite its effectiveness a leader follower and in extension formation control type architecture is inherently limited by node loss seeing mitigated effectiveness, poor collective decision making and diminished adaptability to dynamic environments. In contrast to a centralised formation control, a distributed network swarm can better respond to both dynamic and static environments due to its reliability, flexibility and robust nature \cite{yang2013stability}. Such a swarm architecture sees the swarm base its decision making on the information derived from nearby agents and is able to guarantee the overall swarm stability during maneuvering while avoiding obstacles inside the swarm.\\

\subsection{Obstacle Avoidance Measures}
\label{lit:obstacle_avoid}
A drone swarm requires the ability to dynamically pathfind to its target to avoid both static and dynamic obstacles in its path. A failure to do so will see the swarm fail to achieve its objective and see drones get damaged or destroyed. Initial models to achieve this was done through the generation of Artificial Potential Flow (APF) \cite{APF1} fields to repel the drones from obstacles by generating a flow source. This model is often highly criticised for its ability to create stagnation points potentially locking the drones in place, however this only occurs if a a sink acting as a target exists in the model. Alternatively due to this limitation Reynolds flocking distribution behavioural model \cite{reynolds1987flocks} and Particle swarm optimization \cite{PSO}  were introduced to provide a more reliable and robust method to path plan to targets around obstacles. These methods allow for more complex path planning around obstacles improving efficiency and convergence times but do so with significantly higher computational demand. However despite the advancements made in this methodologies, in our case a APF obstacle as a flow source is effective due to us not designating our target as a flow sink removing any chance of stagnation, this is due to the model being the simplest its ability to be easily expanded to any shape.  

\subsection{Dynamic Trajectory Prediction Methodology}
\label{lit:predictive_models}
Trajectory prediction was initially achieved through Long Short Term Memory (LSTM) and Curve fitting approximations which have been continually refined through various configurations and optimizations to remain as current state of the art methods. Principally LSTM has been used extensively in spacial temporal trajectory prediction either through tracking and predicative relative positions \cite{LTSM1} or by directly tracking and analysing their coordinates in the spacial plane \cite{LSTM2}. However due to LSTMs back propagation in its learning and adaption laws its significantly slower than other means and requires extensive training on data sets to make predictions with some complex models requireing thousands of unique sets to refine its prediction \cite{LSTM6000}.\\

Alternatively curve fitting can be utilized for rapid tracking and learning however curve fitting has to balance the dynamic between using high order polynomials for increased accuracy at the expense of slow computational times or low order polynomials with quick computation. Consequently its often utilized as a supporting element within larger deep learning models \cite{vessel_trajectory_curvefit} to enhance their performance. These models have the ability when utilized correctly to provide high degrees of accuracy in their prediction, however no method is currently capable of learning online without any prior training. \\

However Bidirectional Brain Emotional learning is an adaptive Fuzzy Neural Network controller that has shown excellent online adaptability to systems this has been demonstrated through real time adaptive control systems for quad copters \cite{BFBEL_adapt_quad} showing excellent performance as a stabilizer to disturbances and through \cite{33gNanoDrone} further reinforcing its resilience as a flight computer. Despite this its never been expanded for use as a predictive tool but has been highlighted as possible future work. Consequently this paper aims to combine features of BFBEL and Curve fitting to realise this predictive capability by enhancing the real time adaptability and online earning ability of BFBEL with the quick computation of low order curve fitting to facilitate a new online capable predictive model.\\

\subsection{Data Fusion Concepts}
\label{lit:data_fusion}
Data fusion and multi threading are relatively new endeavours in the world of rapid computing, however it has been proven to dramatically reduce computation times \cite{multithread_advantage}. Critically with a drone swarm with each drone having individual sensors a fusion approach is critical to maximize the performance of the swarm as using data from a single source is extremely vulnerable to information inconsistency and can lack sufficient data to analyse in the short term \cite{Data_fusion_1}. Additionally comparing advanced data fusion algorithms including, Kalman Filters, particle filter, linear minimum mean, large pyramid, wavelet, Neural Networks, Bayesian inference, Dempster-Shafter theory, Fuzzy set theory, rough set through and random set theory \cite{DataFusion2} showed that fundamentally all methods provide adequate fusion of information some performing better for various data sets based on their make. However while these methods are excellent, due to our requirement to have these fusion methodology run at extremely high speeds on comparable small processes (drone chips) this thesis will use a rolling average to fuse them together. Despite this comparably simpler approach \cite{DataFusion2} remarks "there is no one-size-fits-all solution" and the act of any fusion algorithm present provides the greatest benefit with more complex methods seeing negligible gains for high comparative costs. \\

This review has explored use of Single vs Multi Drone systems, formation vs swarm control, obstacle avoidance measures, predictive methodologies subsection, and data fusion processes.The review finds justified the use of use multi drone distributed control structures
with APF driven obstacle avoidance driven by novel Artificial
Intelligence trajectory prediction encompassing rolling average
data fusion approaches. This the sees the swarm utilize state of
the art approaches and novel architecture for the development
of swarm capable for use in a combat zone

\vfill
\section{Background}
\subsection{Bidirectional Fuzzy Brain Emotional Learning Algorithm}
\label{back_BFBEL}
The control scheme \autoref{BFBEL-P_Flight_Computer)Design} is an extension to the Fuzzy Neural Network (FNN) known as Bidirectional Brain Emotional Learning (BBEL), which has rapid adaptability in controlling nonlinear systems. The FNN is aligned with a number of fuzzy membership functions linked to inputs with its weight adaption depending on the strength  of the connected fuzzy layer as demonstrated in \autoref{fig:Flight_Computer}. Each fuzzy layer works to facilitate a range defined though Gaussian distribution to the anticipated boundaries of the system \cite{Collaborativegassourcelocalization}. The system takes an input series \textbf{i} representing the system being analysed and outputs a response $u_{BFBEL}$.\\

\subsubsection{BFBEL-P Network}
\begin{figure}[H]
\includegraphics[width=3.4in]{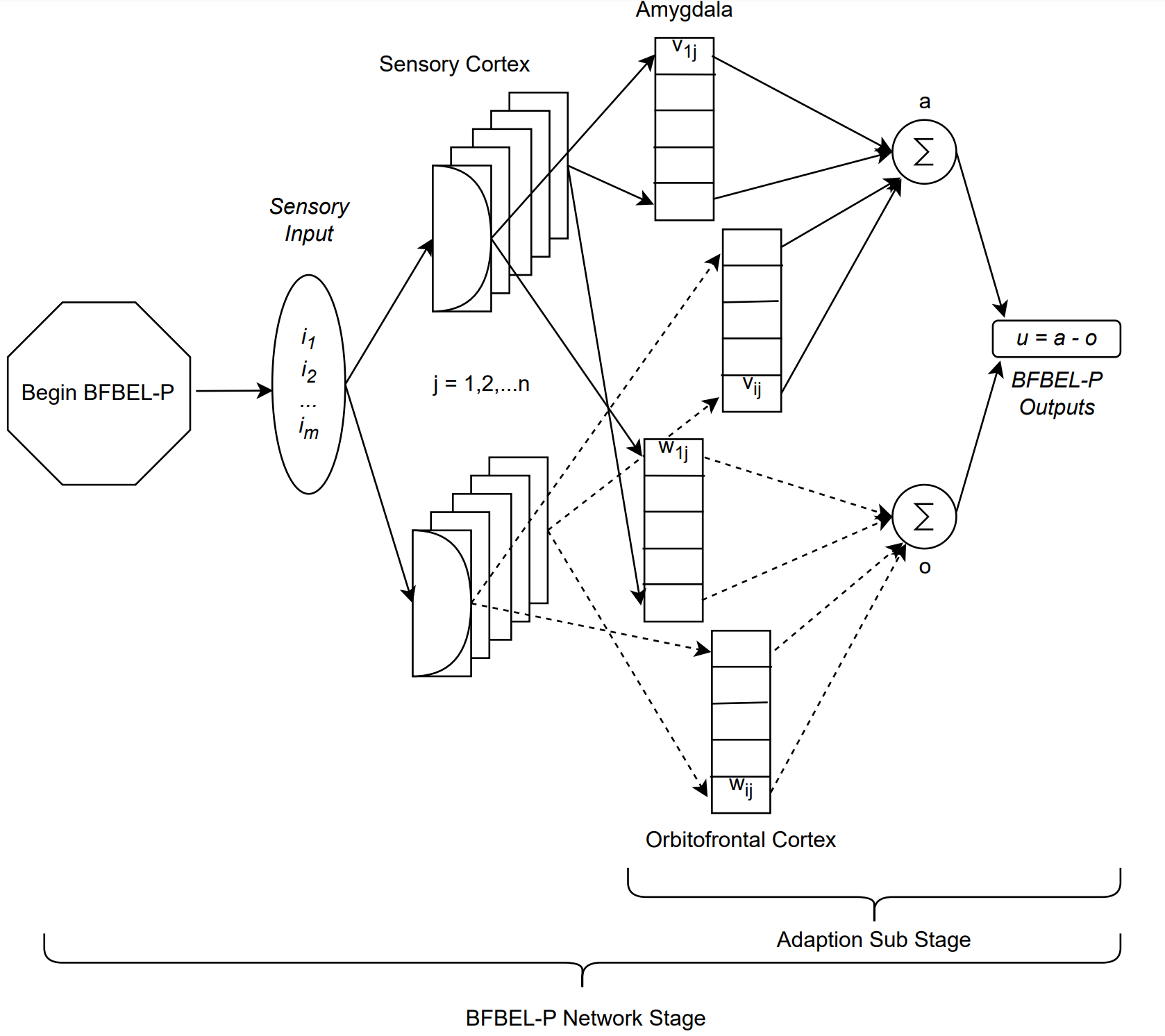}
\caption{BFBEL-P Network Structure.}
\label{fig:BFBEL-P}
\end{figure}

The BFBEL-P Network has 4 distinct sub-stages \cite{33gNanoDrone}, that being the sensory input, sensory cortex, BFBEL network with the Amygdala and Orbitofrontal cortex weighted networks, and finally a BFBEL Output as demonstrated in\autoref{fig:BFBEL-P}.
 The sensory input stage receives a series input as $\mathbf{i} = \begin{bmatrix} i_1,&i_2,&i_3,&\cdots&i_m\end{bmatrix} \in \mathbb{R}$. These inputs flow to the sensory cortex stage where they are mapped through a fuzzy Gaussian membership function to smooth the inputs. The membership function in question is given by $p_{ij} = \exp\left( -\frac{(i_i - \varsigma_{ij})^2}{2\sigma_{ij}^2} \right) \quad \text{where } i = 1, 2, \ldots, m \text{ and } j = 1, 2, \ldots, n.$ Specifically  $\varsigma_{ij}$ represents mean and $\sigma_{ij}$ represents the variance where m and n represent the number of inputs and fuzzy layers respectively. This sees multiple Gaussian fuzzy membership functions (5 in our case) with different ranges overlapping and layered with each input connected to multiple fuzzy layers. These inputs are processed by the amygdala and orbitofrontal cortex multiplying the fuzzy membership functions with their membership weight functions as given by $v_{ij}$ and $w_{ij}$ for the networks respectively. These values are finally summed and their differences subtracted to give the controller output, as formulated by \autoref{eq:u_bfbel_eqn}
\begin{equation}
    \label{eq:u_bfbel_eqn}
    a_i = \sum_{j=1}^{n} p_{ij} v_{ij}, \quad o_i = \sum_{j=1}^{n} p_{ij} w_{ij}, \quad u_{\text{BFBEL}, i} = a_i - o_i
\end{equation}\\

\subsubsection{BFBEL Adaption Laws}
The adaption of Amygdala and Orbitofrontal weights are based upon real mammal brain activity, mimicking the emotional regulation regions of the brain. Specifically the adaption of the $v_{ij}$ and $w_{ij}$ are given as \autoref{eq:adaption_laws}:
\begin{equation}
    \begin{split}
        \label{eq:adaption_laws}
        \Delta v_{ij} = \alpha \left[ p_{ij}(R - a) \right] \\
        \Delta w_{ij} = \beta \left[ p_{ij}(u_{BFBEL} - R) \right]
    \end{split}
\end{equation}

In which $\alpha$ and $\beta$ are the learning rates of the Amygdala and Orbitofrontal Cortex weights respectively. Each weight is updated by summing previous weight to the calculated delta as by \autoref{eq:weight_update}:
\begin{equation}
    \begin{split}
       \label{eq:weight_update}
        v_{ij}(t+1) = v_{ij}(t) + \Delta v_{ij} \\
        w_{ij}(t+1) = w_{ij}(t) + \Delta w_{ij} 
    \end{split}
\end{equation}
Finally the reward signal \autoref{eq:reward_function} is based upon summing the product of the gain vector $q$ and the calculated error and the product of the gain value of the reward signal $c$ and the BFBEL Output:
\begin{equation}
    \begin{split}
        \label{eq:reward_function}
        R = q[error] + c[u_{BFBEL}]
    \end{split}
\end{equation}

\subsection{Swarming Logic}
\label{Swarm_locking_background}
The swarming logic uses Reynolds flocking algorithm \cite{reynolds1987flocks} which sees each boid (UAVs in our case) obey three rules, those being Coherence, Separation and Alignment. Coherence sees the boids steer towards each neighbouring boid at a specified rate, separation is the distance each boid will keep clear around itself from other boids, and finally alignment which sees boids match the vector of the neighbouring boids. This sees boids display a basic flocking behaviour mimicking that of birds and is defined by \autoref{eq:desired_drone_responseX}, \autoref{eq:desired_drone_responseY} and \autoref{eq:desired_drone_responseZ}.
\begin{equation} 
    \label{eq:desired_drone_responseX}
    \Delta x = [x_s - x]R_C + [x - x_o]R_S + [x_s - x_t]R_A
\end{equation}
\begin{equation} 
    \label{eq:desired_drone_responseY}
    \Delta y = [y_s - y]R_C + [y - y_o]R_S + [y_s - y_t]R_A 
\end{equation}
\begin{equation} 
    \label{eq:desired_drone_responseZ}
    \Delta z = [z_s - z]R_C + [z - z_o]R_S + [z_s - z_t]R_A 
\end{equation}
This architecture forms the baseline for all swarming logic and will serve as a backbone implementation for this paper.

\section{Novel Contribution}
This work sees the refinement and enhancement of the BFBEL AI model, previously this model has been used successfully for real time control systems in online settings with no training\cite{33gNanoDrone}. The model showed excellent adaptability, resilience and capability in online learning environments. Currently there is a lack of online capable artificial intelligence models as most require tailor made solutions with large training data sets to achieve predictive capability. However the successful early work on BFBEL showed that there was potential for it to be enhanced to allow future prediction in rapidly changing dynamic environments from its strong adaptability. My work realises this capability and developed the model to allow this, practically this was used in a UAV drone swarm to dynamically track a moving target and maneuver the drones to the target by predicting its forward motion.

\section{BFBEL-P Flight Computer Design}
\label{BFBEL-P_Flight_Computer)Design}
\begin{figure}[H]
\begin{centering}
\includegraphics[width=3.4in]{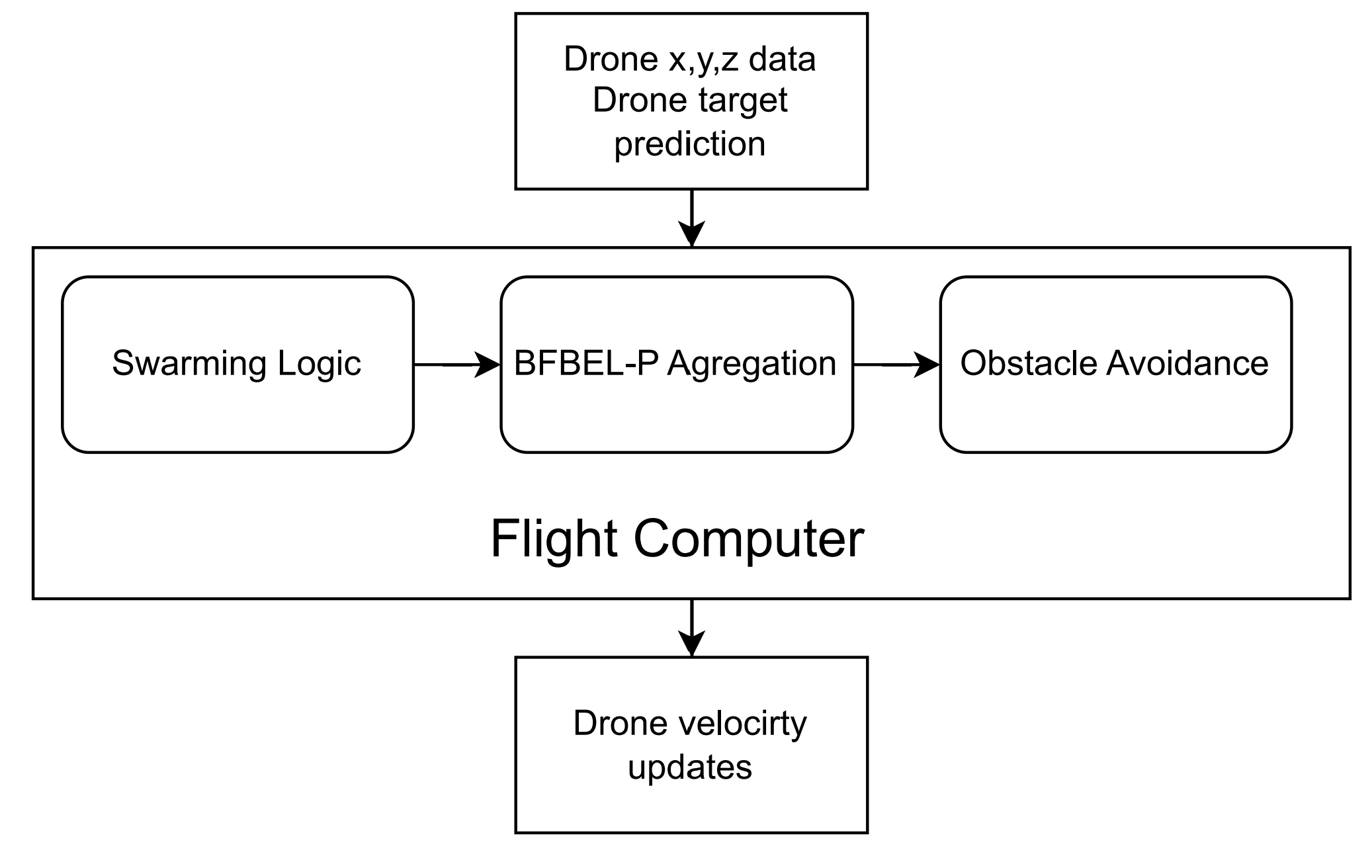}
\caption{Flight Computer Structure.}
\label{fig:Flight_Computer}
\end{centering}
\end{figure}

This work extends on previous research by adapting the BFBEL Network detailed in \autoref{back_BFBEL} to enable future predictions on a system that adapting to in real time without prior training. This system is designated as Bidirectional Fuzzy Brain Emotional Learning Predictive (BFBEL-P) which works through the analysis of the FNN response outputs \autoref{eq:u_bfbel_eqn} over a interval and using these over multiple simultaneous runs to generate artificially feedback to enable the controller to adapt to the hypothetical system enabling a prediction. The mathematical equations and models defining this process is detailed in \autoref{BFBEL-P_Flight_Computer)Design}. The overall control architecture sees each drone independently run a BFBEL-P Prediction which sees a unique solution for every drone in the swarm which are aggregated to produce a final target prediction. This is achieved through a flight computer depicted in \autoref{fig:Flight_Computer} which receives every drone prediction along with their inertial measurement unit (IMU) data. This passes through 3 primary components, that being the swarming logic, the BFBEL-P Aggregation and the obstacle avoidance measures which then publishes velocity commands to every drone within the swarm who use PID control methodologies to adjust the drones parameters to the received commanded which is detailed in \autoref{PID Control}.

\subsection{Swarming Logic}
The drones swarm using Reynolds flocking as detailed in \autoref{Swarm_locking_background}. This aggregates the location and the IMU data of all drones within a pre-determined visual range to generate a swarm. The computer then uses this logic to publish velocity commands to each drone to keep them swarmed within the tolerances allowed. Critically this allows of multiple swarms to run simultaneously within the same set of drones with seamless merging and splitting of the swarm into sub swarms should the operational demand be there to avoid obstacles.

\subsection{BFBEL-P Structure}
\begin{figure}[H]
\includegraphics[width=3.4in]{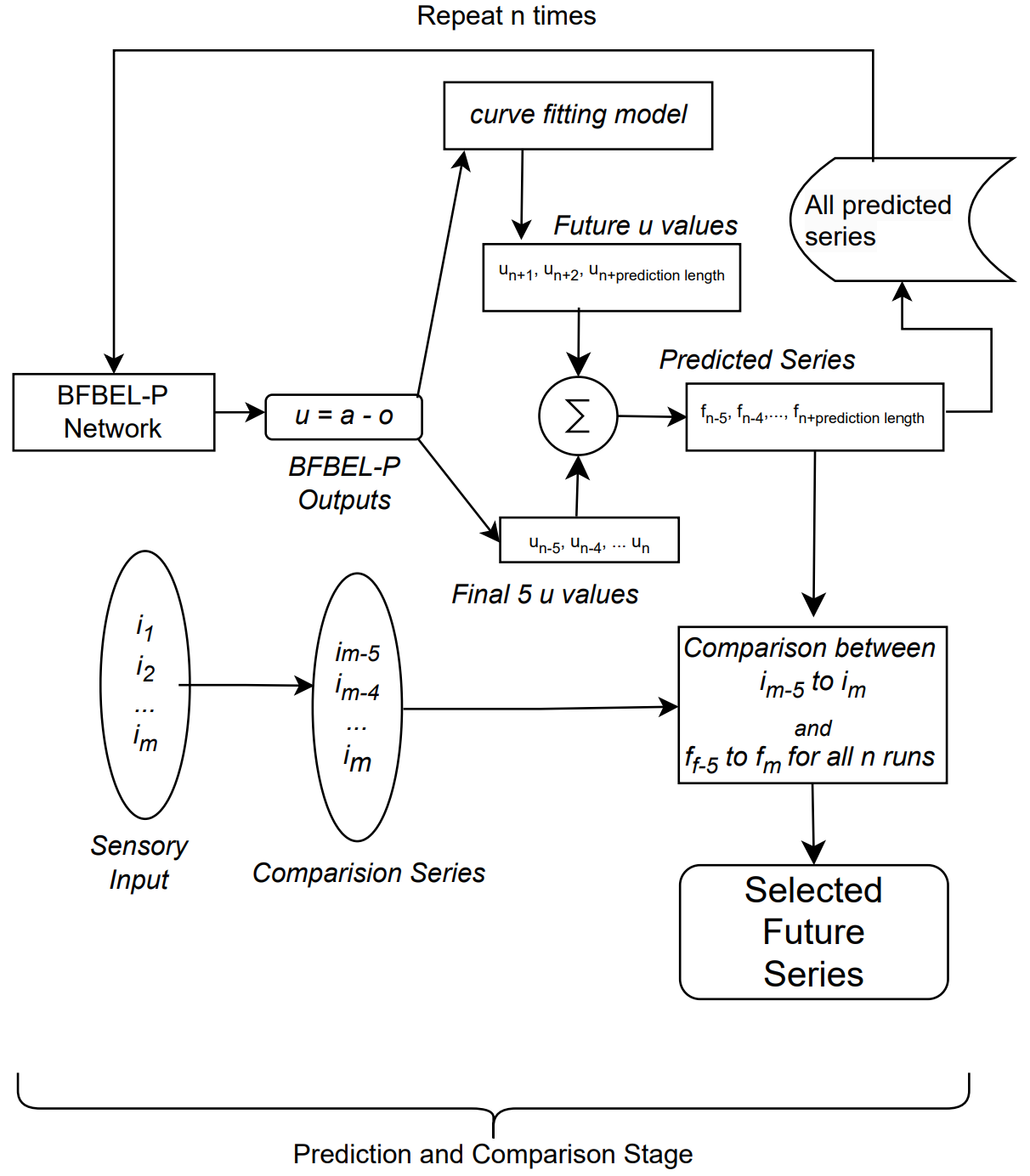}
\caption{Flight Computer Prediction and Comparison Stage.}
\label{fig:prediction_comparison}
\end{figure}

The BFBEL-P Structure which is run on every drone is separated into two primary components, that being the BFBEL-P Network itself including its adaption laws specifically \autoref{eq:adaption_laws}, \autoref{eq:reward_function}, \autoref{eq:error_function} and the predictive and comparison stage shown in \autoref{fig:prediction_comparison}. The structures receives the sensory target data and outputs to the flight computer a unique prediction for every drone in the swarm.\\

\subsubsection{BFBEL-P Prediction and Comparison }
The model takes the target coordinates in $t$ iteration ($x_t, y_t, z_t$) as the input vector $\textbf{i}$ to the BFBEL-P network, while the output vector $u_{BFBEL}$ in series which are stored in long-term array. Critically the BFBEL-P network is adapted based upon the error \autoref{eq:error_function} which is determined by the difference between the predicted value and the actual value of the given input sequence during the learning process:

\begin{equation}
    \begin{split}
        \label{eq:error_function}
        error = (i_{i} + u_{BFBEL}) - i_{i+1}
    \end{split}
\end{equation} 

Fitting the $u_{BFBEL}$ values on a third order curve fitting approximation given by the polynomial: $ax^3 + bx^2 + cx + d$ This enables the $u_{BFBEL}$ values to be predicted into the future and stored as a form of artificial feedback for the controller to stabilize it, using the predicted $u_{BFBEL}$ values we can generate a prediction of values from the end of the sensory input stage. However to enable an effective comparison and prediction selection process this predictions starts at $i_{-5}$ and goes to the desired future prediction length as demonstrated in
\(\mathbf{f} = \begin{bmatrix} f_{m-5}, & f_{m-4}, & \cdots & f_{m+\text{prediction length}} \end{bmatrix}\).
This is then compared to the final true 5 values of the sensory input given by
\(\mathbf{c} = \begin{bmatrix} i_{m-5}, & i_{m-4}, & \cdots & i_{m} \end{bmatrix}\).
The BFBEL controller runs this predictive stage multiple times to generate multiple \(\mathbf{f}\) predictions which can be filtered into the best future prediction. The filtering process sees a comparison between the \(\mathbf{f}\) series and the comparison series \(\mathbf{c}\), critically analyzing the sum of differences between true and predicted values \autoref{eq:Sum_of_Differences}, the error difference in their slopes \autoref{eq:diff_in_slopes}, and the error in their curvature \autoref{eq:diff_in_curvature}, giving a net error between the final segment of true values and the first segment of predicted values.
\autoref{eq:sum_of_errors}.
\begin{equation} 
    \label{eq:Sum_of_Differences}
    e1 = \sum(\mathbf{f} - \mathbf{c})^2
\end{equation}

\begin{equation} 
    \label{eq:diff_in_slopes}
    e2 = sum (\frac{\partial \mathbf{f}}{\partial x} - \frac{\partial \mathbf{c}}{\partial x}) + (\frac{\partial \mathbf{f}}{\partial y} - \frac{\partial \mathbf{c}}{\partial y})
\end{equation}

\begin{equation}
    \label{eq:diff_in_curvature}
    e3 = \frac{| \frac{\partial^2 \mathbf{f}}{\partial x^2} \frac{\partial \mathbf{f}}{\partial y} - \frac{\partial^2 \mathbf{f}}{\partial y^2} \frac{\partial \mathbf{f}}{\partial x} |}{\left( \frac{\partial \mathbf{f}}{\partial x}^2 + \frac{\partial \mathbf{f}}{\partial y}^2 \right)^{3/2}} - \frac{| \frac{\partial^2 \mathbf{c}}{\partial x^2} \frac{\partial \mathbf{c}}{\partial y} - \frac{\partial^2 \mathbf{c}}{\partial y^2} \frac{\partial \mathbf{c}}{\partial x} |}{\left( \frac{\partial \mathbf{c}}{\partial x}^2 + \frac{\partial \mathbf{c}}{\partial y}^2 \right)^{3/2}}
\end{equation}

\begin{equation} 
    \label{eq:sum_of_errors}
    comparison\_error = e1 + e2+ e3
\end{equation}
The error is then compared against all \textbf{f} series and that with the lowest error is selected as the best prediction to be fed into the flight computer. This process works to filter out a degree of noise within the computer that is present from the BFBEL-P's learning. Principally the predictive model is highly susceptible to the learning trends of the controller generating this requirement. 

\subsection{Prediction Aggregation}
\begin{figure}[H]
\includegraphics[width=3.4in]{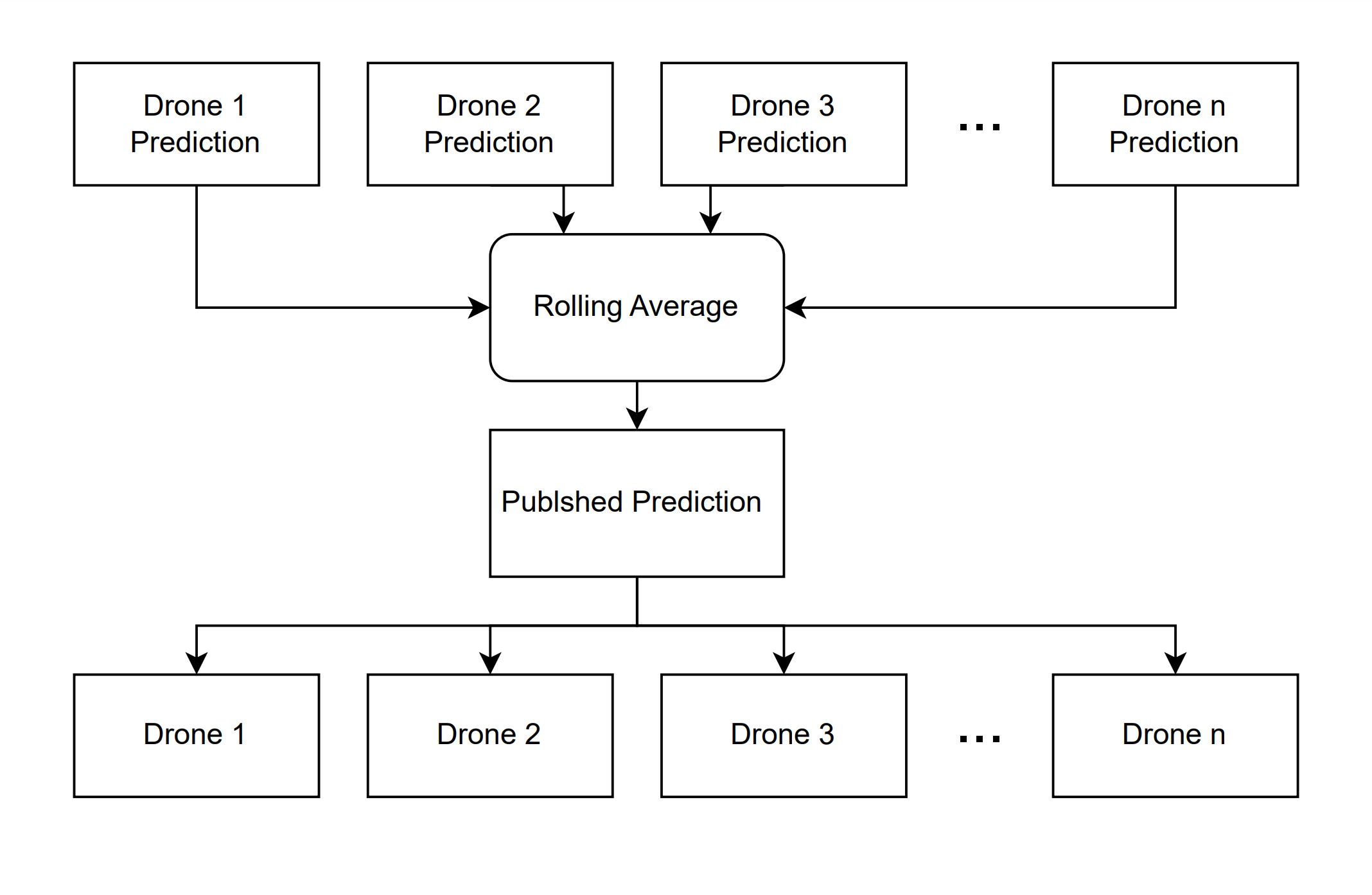}
\caption{Prediction Aggregation Data Flow.}
\label{fig:prediction_agregation}
\end{figure}

To achieve high convergence times for target prediction, every drone in the swarm generates a unique prediction which is aggregated by the flight computer in a rolling average \autoref{fig:prediction_agregation}.
Specifically the rolling average can be defined through \autoref{eq:rollin_avearge}
\begin{equation} 
    \label{eq:rollin_avearge}
    \bar{x}_n = \frac{x_n + x_{n-1} + \dots + x_{n-(N-1)}}{N}
\end{equation}
Critically $\bar{x}_n$ is equated at each iteration of the fight computer with all predictions completed during this time stored in memory awaiting the computation. This allows the flight computer to still compute the solution even in the event of drones predictions failing running slow increasing the robustness of the system. Consequently compared to a centralized prediction, the trials and computation can be distributed overall of the drones seeing the computation time decrease linearly with the addition of more drones. Furthermore within the ROS simulation there is a degree of randomness in the suit, principally that individual drones can run slightly faster or slower than at their desired 30hz rate ($<$5 \% variation). However despite this the controller has no requirement to receive information from every drone to aggregate a net solution but rather will take as many as available to produce the strongest results, furthermore the computer is able to still function with multiple independent swarms working together as demonstrated within \autoref{Linear Target Performance Analysis} consequently demonstrating the robustness of the system. Furthermore the average disregards solutions that are more than 2 standard deviations above the mean however with conditions in place to allow it if multiple drones create similar predictions. This stabilizes the prediction leading to improved overall accuracy in prediction as shown in \autoref{fig:Linear_BFBELP_Analysis} and \autoref{fig:fig8_BFBELP_Analysis}.

\subsection{Obstacle Avoidance}
Obstacle avoidance is achieved through the generation of a artificial potential flow (APF) field around a designated point defined through \autoref{eq:APF_Field} from \cite{APF1}. 
\begin{equation} 
    \label{eq:APF_Field}
    U_{\text{rep}}(x, y) = 
    \begin{cases} 
\frac{1}{2} \eta \left( \frac{1}{\rho(x, y)} - \frac{1}{\rho_0} \right)^2 & \text{if } \rho(x, y) \leq \rho_0 \\
0 & \text{if } \rho(x, y) > \rho_0
\end{cases}
\end{equation}
This allows the drones to be pushed away from obstacles whilst reaching its target to demonstrate the robustness of the controller, however this can be enhanced in future work through the implementation of LIDAR to navigate more complex terrain which can be mapped with APF fields. 

\section{Methodology}
The simulations were run on a native Ubuntu 20.04 environment using ROS Noetic with a Gazebo simulation environment to simulate real world physics and effects. This was driven by a 2019 Microsoft Surface Book 2 running an Intel Core i7-8650U CPU @ 190GHzm 8GB of DDR4 RAM and a NVIDIA GeForce GTX 1050 GPU. The drones were generated using the Ardrone package simulating the Parrot AR.Drone 2.0 UAV which is equivalent to the first person view (FPV) drones currently being used in Ukraine. To define the motion of the target a custom target script is produced with the desired geometry (figure 8 in our case) which is being published to be read by the drones, furthermore the obstacles and their nature can be defined within the same processes on a custom obstacle script. The flight computer then reads this published information and subscribes to the IMU and target predictions from all the drones located within a swarm before doing its internal calculations and publishing commands to the drones as demonstrated in the ROS Quick Toolkit plot (rqt plot) in \autoref{fig:rqt_plot}, the data pathways are further explained and visualized in \autoref{RQT Plot Information}. Within ROS two simulation environments are defined for analysis based upon the targets motion, which each get analysed in 10 identical runs for analysis in MATLAB post simulation. 

\section{Simulation Results and Discussion}
The system was tested in two environments, that being against a target moving in linear line given by $y = t, x = t$ and by a target moving in a figure eight defined by $y = 50sin(\frac{2\pi}{200}t), x = 50sin(\frac{2\pi}{200}t)$ where $t$ is the simulation time both with an obstacle located on their paths at (30, 30) and (10, 10) respectively. 

\subsection{Linear Simulation}
The purpose of the linear target simulation case is to test the fundamental swarming, target tracking and obstacle avoidance capabilities in a simple setting with more controlled variables. A full linear trial is analysed as demonstration in \autoref{Linear Target Performance Analysis}. However analysing the the performance across all trials sees an average confidence interval of [-0.57176, 0.25072] in the X domain and [-0.55727, 0.25700] in the Y domain with 95\% confidence and all trials found to have statistically different average predictions by virtue of a T test. Furthermore averaging the group metric before being influenced by a obstacle (t $<$ 25s in \autoref{fig:Linear_Group_Metric}) sees an average of 1.7m separation, which is comfortably above the 1.5m minimum and remains within its acceptance threshold, moreover the order metric \autoref{fig:Linear_Order_Metric} sees on average 5.6s with zero angular difference with peaks at 0.29 rad/s within the same time interval. The combination of these two metrics demonstrate excellent swarming performance in ideal cases, verifying the capabilities of the swarming logic. Each trails also saw no collisions, and all tracked to the target however noting that the swarm cohesion post the obstacle saw increased instability due to overturned responses which could be improved by additional response dampening. That being said the the drones were able to track to the moving linear target which was at times obscured by an obstacle with a strong confidence interval indicating strong fundamental swarming, predictive and obstacle avoidance capabilities achieving the outcomes of the test.

\subsection{Figure Eight Simulation}
The purpose of the figure eight simulation case is to test the full extent of swarming, target tracking and obstacle avoidance capabilities in a complex scenario attempting to replicate a maneuvering target in a war zone. A full figure eight trial is analysed as demonstrated in \autoref{Figure Eight Target Performance Analysis}. However analysing the the performance across all trials sees an average confidence interval of [-0.44713, 0.30497] in the X domain and [-49754, 0.56195] in the Y domain with 95\% confidence and all trials found to have statistically different average predictions by virtue of a T test. Furthermore analysing the group metric before the first turn (t $<$ 80s in \autoref{fig:fig8_Group_Metric}) sees the average separation fluctuate sinusoidal from 2.5 to 1m this is indicative and aligns with the delayed Y domain responses as detailed in \autoref{Figure Eight Target Performance Analysis}, combined with strong X and Z domain performance. This is identified through the sinusoidal group metric aligning with the corresponding domain error plots. Despite this from the linear trials its identified that the swarming performance is strong further reinforcing the notion that the separation distance variation is due to Y domain responses. Moreover the order metric \autoref{fig:fig8_Order_Metric} sees fluctuation between 0 and 0.275 rad/s average angular velocity difference which is indicative of the swarm struggling to match the rapid Y domain motion of the target at the peak of the turn. However the swarm shows zero difference in the early stages of its flight demonstrating excellent fundamental swarming behaviour that required some additional tuning to the minimums and maximal limits of the drones motion to enable faster tracking. All trails also saw no collisions, and all tracked to the target. This test achieves its objectives by finding the limits of the drone swarm performance however despite the challenging trial it still demonstrated excellent predictive and target tracking capabilities which is slightly overshadowed by performance in one domain, dominated by PID response tuning.

\subsection{BFBEL-P Performance Analysis}
Analysing the BFBEL-P performance across all runs sees that the prediction falls on average within confidence interval of [-0.57176, 0.25072] in the X domain and within [-0.55727, 0.25700] for the Y domain with 95\% confidence across the linear simulation and within [-0.44713, 0.30497] for the X domain and within [-0.49754, 0.56195] for the Y domain for the figure eight simulation. Additionally T tests were conducted for all trials with all 10 linear trials found to have predictions statistically similar to the Targets and with 9 of 10 figure eight simulations being statistically similar. Consequently we can explicitly determine that the BFBEL-P prediction model is accurate and valid for predictive purposes in both simple and complex trajectory. To further analyse the performance of BFBEL-P, it was compared to other state of the art methods in its predictive capability in both short and long term cases. \\

\subsubsection{Short Term Testing}
The first test for BFBEL-P sees it track a figure eight moving on a rapid time interval seeing a total of 90 predictions. The purpose of this test is to see its adaptability compared to other state of the art methods which is demonstrated in \autoref{fig:BFBEL_Pred_comparison}, \autoref{tab:Numerical_Comp_BFBEL} and \autoref{tab:Numerical_interval_short_BFBEL}

\begin{table}[H]
\begin{center}
\caption{BFBEL-P, Curve Fitting and LSTM Predictive Short Term Time Comparison}
\label{tab:Numerical_Comp_BFBEL}
\begin{tabular}{| p{3.8cm} | p{2cm} |}
\hline
Prediction Method& Time Taken Per Prediction (s)\\
\hline
BFBEL Time Single Drone& 1.195710203
\\ 
\hline
BFBEL 4 Drones & 0.298927551
\\ 
\hline
Curve Fitting& 
0.002519895\\ 
\hline
 LSTM &
0.318216667\\\hline
\end{tabular}
\end{center}
\end{table}

\begin{table}[H]
\begin{center}
\caption{BFBEL-P, Curve Fitting and LSTM Predictive Short Term Confidence Interval Comparison}
\label{tab:Numerical_interval_short_BFBEL}
\begin{tabular}{| p{3.4cm} | p{2.4cm} |}
\hline
Prediction Method& Confidence Interval\\
\hline
BFBEL-P X Domain & [2.503, 4.410]
\\ 
\hline
BFBEL-P Y Domain & [2.999, 5.337]
\\ 
\hline
Curve Fitting X Domain & [0.194, 0.915]
\\ 
\hline
Curve Fitting Y Domain & [9.774, 13.016]
\\
\hline
LSTM X Domain & [0.162, 1.206]
\\
\hline
LSTM Y Domain & [1.874, 0.926]
\\\hline
\end{tabular}
\end{center}
\end{table}

These tests show that BFBEL-P has strong adaptability to a rapidly moving target and demonstrating strong comparative performance to alternate state of the art methods whilst still occurring in a sound time frame, noting that the time taken will drop with having more drones in the swarm.\\

\subsubsection{Long Term Testing}
The second test for BFBEL-P sees it track a figure eight moving on a slow time interval seeing a total of 840 predictions. The purpose of this test is to see its long term accuracy compared to other state of the art methods which is demonstrated in \autoref{fig:long_term_comparison}, \autoref{tab:Numerical_Comp_BFBEL_Long} and in \autoref{tab:Numerical_interval_long_BFBEL}

\begin{table}[H]
\begin{center}
\caption{BFBEL-P, Curve Fitting and LSTM Predictive Short Term Time Comparison}
\label{tab:Numerical_Comp_BFBEL_Long}
\begin{tabular}{| p{3.8cm} | p{2cm} |}
\hline
Prediction Method& Time Taken Per Prediction (s)\\
\hline
BFBEL Time Single Drone& 0.57151025
\\ 
\hline
BFBEL 4 Drones & 0.142877562
\\ 
\hline
Curve Fitting& 
0.0014774
\\ 
\hline
 LSTM &
0.291874613\\\hline
\end{tabular}
\end{center}
\end{table}

\begin{table}[H]
\begin{center}
\caption{BFBEL-P, Curve Fitting and LSTM Predictive Long Term Confidence Interval Comparison}
\label{tab:Numerical_interval_long_BFBEL}
\begin{tabular}{| p{3.4cm} | p{2.4cm} |}
\hline
Prediction Method& Confidence Interval\\
\hline
BFBEL-P X Domain & [-2.049, 1.877]
\\ 
\hline
BFBEL-P Y Domain & [-1.987, 1.939]
\\ 
\hline
Curve Fitting X Domain & [-1.909, 2.016]
\\ 
\hline
Curve Fitting Y Domain & [-1.924, 2.001]
\\
\hline
LSTM X Domain & [-2.009, 1.917]
\\
\hline
LSTM Y Domain & [-1.991, 1.934]
\\\hline
\end{tabular}
\end{center}
\end{table}

These second trials show virtually indistinguishable results from comparative methodologies in a substantially quicker time frame to the first test which occurs due to simpler calculations being run by the BFBEL-P computer with a slow maneuvering target. This validates the capabilities of the BFBEL-P Model whilst highlighting its efficiency as a model.

\subsection{Swarming Performance Analysis}
Analysis of the Swarming performance occurs through the group and order metrics, critically during the regions of flight where the swarm has not been disturbed by external influences such as an obstacle, The analysis uses the linear simulation case due to its simplicity, this in tern provides us the best understanding of the swarm performance. Analysing the trials as mentioned previously sees averages 1.7m of separation and sees angular velocity difference peaks at 0.29 rad/s with 25\% of the time spent at 0 before external influences \autoref{fig:Linear_Order_Metric}. However despite this, imperial analysis of the motion such as in \autoref{fig:Linear_3D Plot} and \autoref{fig:fig8_3D Plot} after influence from an obstacle requires some smoothing as its comparably incoherent in the individual drone motion. However despite this volatility the drones stay on mission and critically never collided in the trials. Additionally the drones in both figure eight and linear simulations were placed in scenarios where the Swarm had to split into smaller groups and rejoin, while this demonstrates the robustness of the swarm the rejoining process was longer than desired and relatively chaotic in relation to nominal swarm motion. Subsequently adding a link with longer range between individual swarmed groups could enable them them to collocate earlier and slower to enable a smoother transition whilst not influencing the individual swarms stability.

\section{Conclusion}
This paper introduces a novel predictive model extending on current FNN technology and methods. Emotional learning has significant advantages due to enhanced real time adaptability and learning unlike traditional models which require tuning or prolonged training. This allowing for a generic model to be used for any system in any environment and still provide accurate predictions. Furthermore the model proposes a data fusion strategy with multi threading across drones in the swarm enabling a robust system where computation is shared across the swarm allowing for missions to continue despite drones being lost, furthermore this architecture enables decreased computational times with more drones in the loop. Such a predictive model is ideal for the implementation onto swarms of drones in a military context due to its unpredictable nature and requirement for high flexibility to dynamic circumstances. Our simulations and tests in both simple and complex target motion shows faster convergence improved accuracy in the short term and indistinguishable differences to state of the art methods in the long term. Future work involves further enhancing and tuning of the fundamental swarming logic to improve the flight performance and response to keep pace with predictive capability, furthermore this system can be integrated outside of the simulation environment to real drones to practically validate its performance.


\bibliographystyle{IEEEtran}
\bibliography{references}


\clearpage
\appendices

\section{PID Control Architecture}
\label{PID Control}
A PID controller is a robust closed-loop control algorithm that operates on three coefficients: proportional, integral, and derivative which are used in conjunction to determine the optimum system response. The output of a PID controller is calculated in the time domain from the feedback error as demonstrated in \autoref{PID output u(t)}: 

\begin{equation} 
    \label{PID output u(t)}
    u(t) = K_{p}e(t) + K_{i}\int e(t)dt + K_{d}\frac{de}{dt}
\end{equation}

The proportional $(K_p) $ gain is used to correct an error. The derivative $(K_d) $ gain is used to add damping to a system and improve system response and limit overshoot. Integral $(K_i)$ gain is used to remove steady-state errors but will reduce stability. This is often written in the form of the PID controller transfer function, which is found by taking the Laplace transformation of \autoref{PID output u(t)}, resulting in \autoref{PID Transfer Function}

\begin{equation}
    \label{PID Transfer Function}
    K_{p} + \frac{K_{i}}{s} + K_{d}s = \frac{K_{d}s^{2} + K_{p}s + K_{i}}{s}
\end{equation}

\begin{figure}[H]
\includegraphics[width=3.4in]{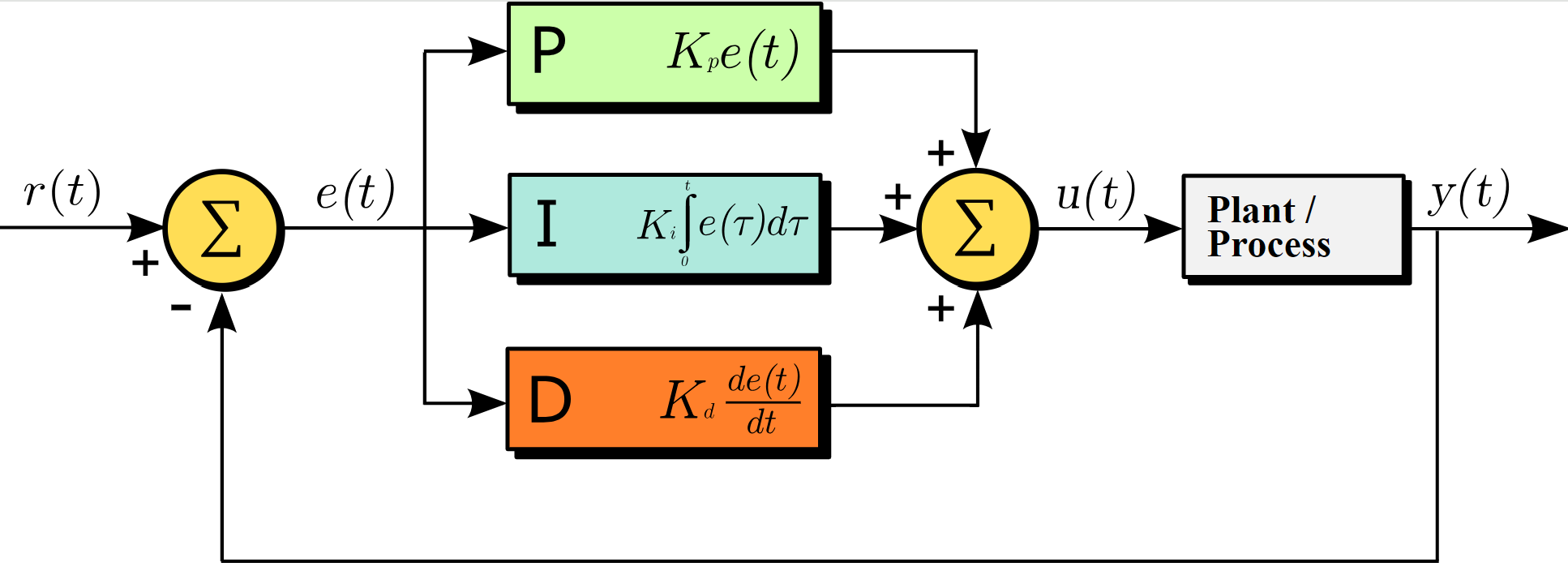}
\caption{PID Control Architecture }
\label{fig:PID_Diagram}
\end{figure}

The control signal $(u)$ from the PID controller drives the aircraft response (Plant/Process), resulting in a new output $(y)$. This output is then compared to the reference $(r)$ to calculate a new error $e$. This feedback loop persists as long as the controller is active. The overall architecture of such a system is depicted in \autoref{fig:PID_Diagram} from \cite{PID_image}.

\section{Linear Target Performance Example}
\label{Linear Target Performance Analysis}
\begin{figure}
\includegraphics[width=3.4in]{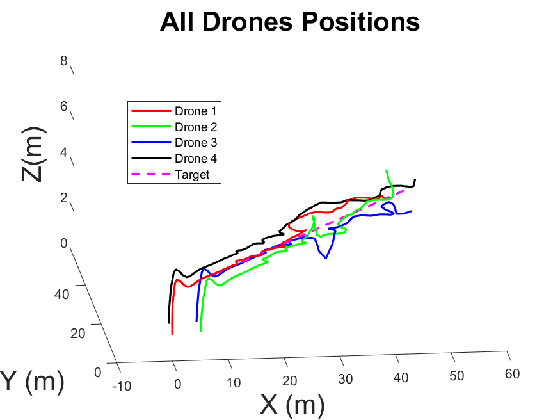}
\caption{Linear Trial 3D Plot}
\label{fig:Linear_3D Plot}
\end{figure}

\begin{figure}
\includegraphics[width=3.4in]{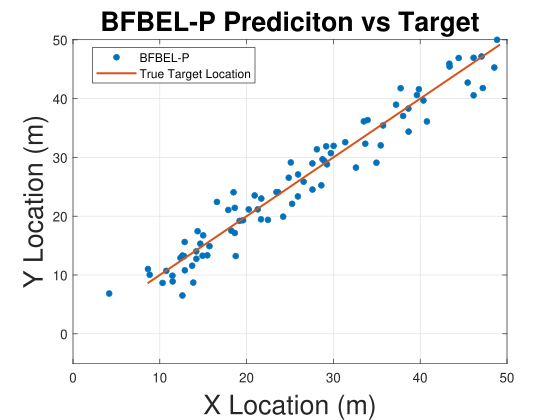}
\caption{Linear Trial BFBEL-P Performance Analysis}
\label{fig:Linear_BFBELP_Analysis}
\end{figure}

\begin{figure}
\includegraphics[width=3.4in]{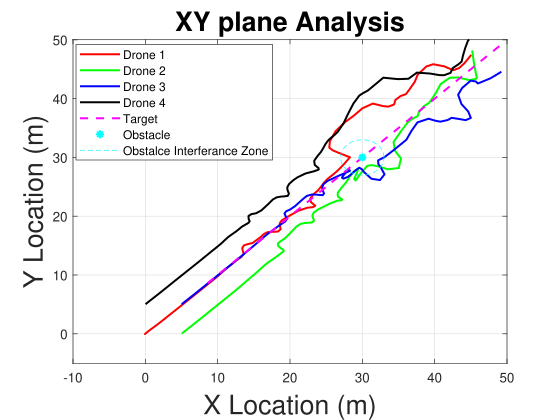}
\caption{Linear Trial xy plane with obstacle analysis}
\label{fig:Linear_xy_analysis}
\end{figure}

\begin{figure}
\includegraphics[width=3.4in]{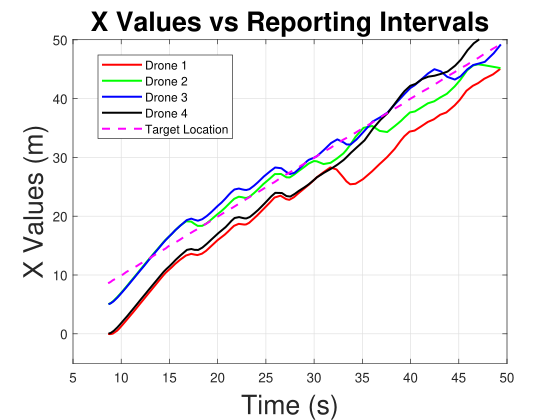}
\caption{Linear Trial x location analysis}
\label{fig:Linear_x_loc}
\end{figure}

\begin{figure}
\includegraphics[width=3.4in]{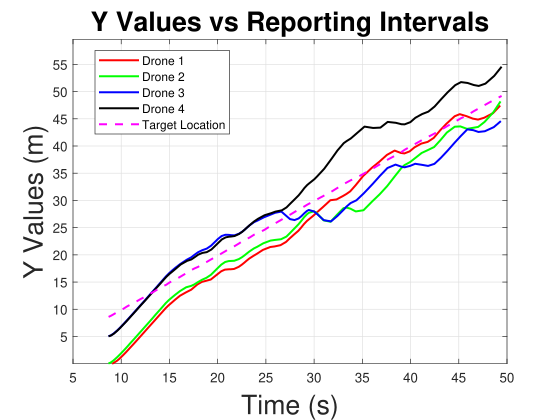}
\caption{Linear Trial y location analysis}
\label{fig:Linear_y_loc}
\end{figure}

\begin{figure}
\includegraphics[width=3.4in]{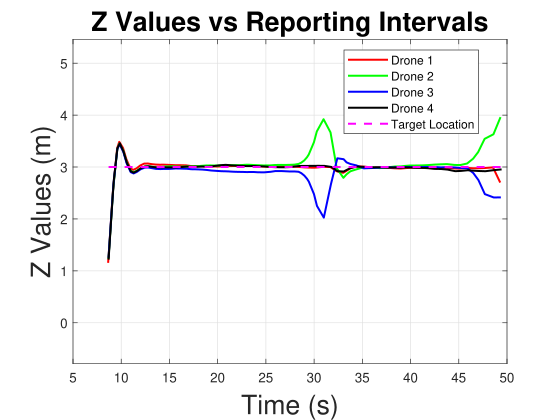}
\caption{Linear Trial z location analysis}
\label{fig:Linear_z_loc}
\end{figure}

\begin{figure}
\includegraphics[width=3.4in]{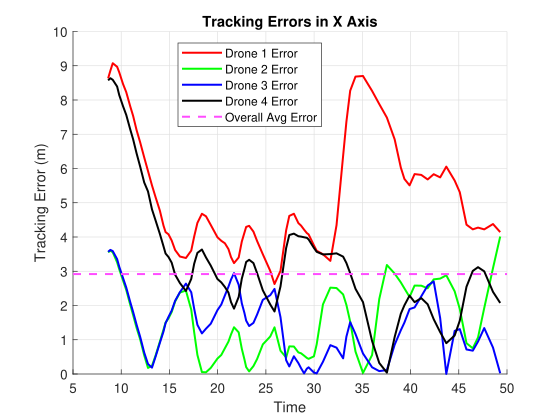}
\caption{Linear Trial x location tracking error}
\label{fig:Linear_x_tracking_error}
\end{figure}

\begin{figure}
\includegraphics[width=3.4in]{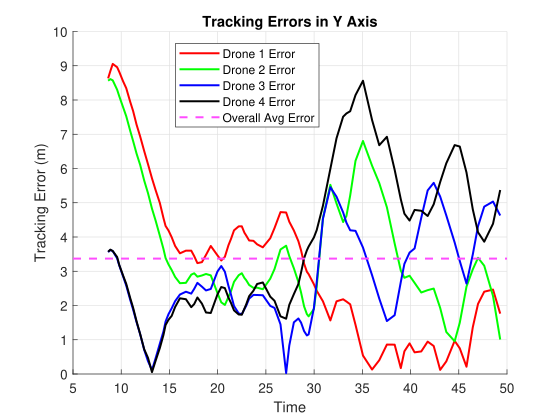}
\caption{Linear Trial y location tracking error}
\label{fig:Linear_y_tracking_error}
\end{figure}

\begin{figure}
\includegraphics[width=3.2in]{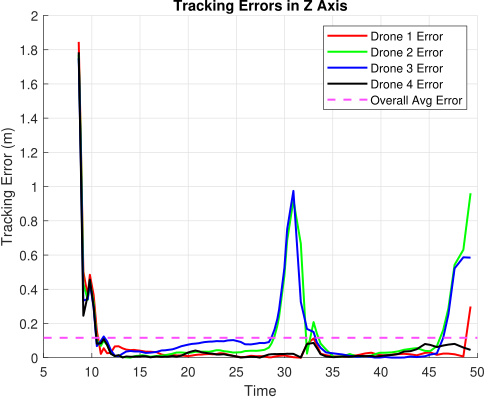}
\caption{Linear Trial z location tracking error}
\label{fig:Linear_z_tracking_error}
\end{figure}

\begin{figure}
\includegraphics[width=3.4in]{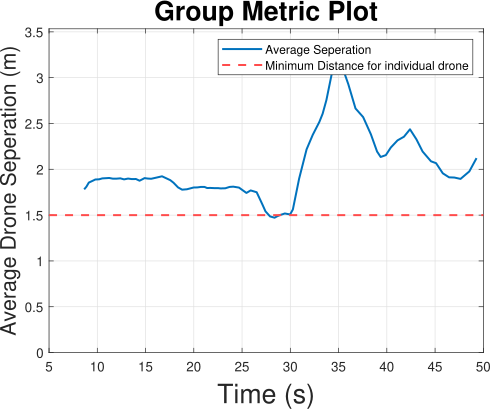}
\caption{Linear Trial Group Metric Plot}
\label{fig:Linear_Group_Metric}
\end{figure}

\begin{figure}
\includegraphics[width=3.4in]{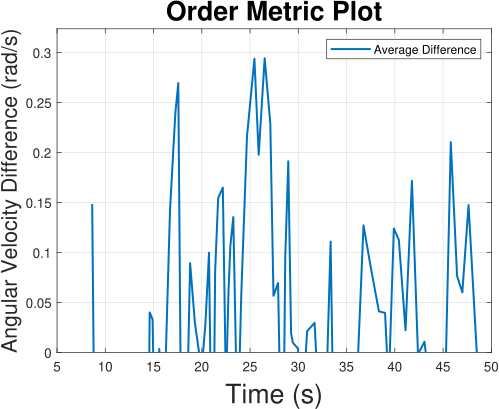}
\caption{Linear Trial Order Metric Plot}
\label{fig:Linear_Order_Metric}
\end{figure}

For each Linear trial the following plots are generated that being a 3D plot \autoref{fig:Linear_3D Plot}, BFBEL-P vs Target Prediction Comparison \autoref{fig:Linear_BFBELP_Analysis}, X Location Comparison \autoref{fig:Linear_x_loc}, Y Location Comparison \autoref{fig:Linear_y_loc}, Z Location Comparison \autoref{fig:Linear_z_loc}, XY Plane plot \autoref{fig:Linear_xy_analysis}, Group Metric \autoref{fig:Linear_Group_Metric}, Order Metric \autoref{fig:Linear_Order_Metric}, X Tracking Error \autoref{fig:Linear_x_tracking_error}, Y Tracking Error \autoref{fig:Linear_y_tracking_error} and Z Tracking Error \autoref{fig:Linear_z_tracking_error}. Additionally for each trial confidence intervals are generated for the BFBEL-P predictions in both the x and y axis which also gets analysed in a T test.  Additionally the average tracking errors are calculated along with the average standard deviation and variance for the drones in the x,y,z location compared to the target, this statistical analysis is given in \autoref{tab:linear_trial_stats}.

\begin{table}[H]
\begin{center}
\caption{Statistical Analysis from a linear trial}
\label{tab:linear_trial_stats}
\begin{tabular}{| p{3.8cm} | p{3cm} |}
\hline
Statistical Test & Result \\
\hline
BFBEL-P vs Target Confidence interval (95\%) in X Domain & [-0.3997, 0.4896] \\ 
\hline
BFBEL-P vs Target Confidence interval (95\%) in Y Domain & [-0.6120, 0.2580] \\ 
\hline
BFBEL-P vs Target T-Test in X Domain & p = 0.8413 for $\alpha$ = 0.05 \\ 
\hline
BFBEL-P vs Target T-Test in Y Domain & p = 0.4207 for $\alpha$ = 0.05 \\ 
\hline
Tracking Error X Domain & 2.9168 \\ 
\hline
Tracking Error Y Domain & 3.3689 \\ 
\hline
Tracking Error Z Domain & 0.11641 \\ 
\hline
Average Standard Deviation for Overall Motion & 2.0327 \\ 
\hline
Average Variance for Overall Motion & 4.4396 \\ 
\hline
\end{tabular}
\end{center}
\end{table}

\vspace{5mm} 

From the statistical data we can determine that neither the x or y BFBEL-P predictions are statistically significantly different from the target values, which combined with each result predicting within a meter with 95\% confidence, as reinforced by \autoref{fig:Linear_BFBELP_Analysis}. indicates strong predictive performance by the BFBEL-P Model. Additionally it shows sound tracking error in the x and y plane showing  exceptional accuracy in the z axis, however critically noting some variance and standard deviations in the results which are indicative of some instability in the swarms motion. Furthermore analysing the obstacle avoidance in \autoref{fig:Linear_xy_analysis} we can see the drones avoided the target once moved into the interference zone critically noting that the target was located directly on top of where the target was at the time and stayed within the interference zone for a 10s period (x and y motion was defined by simulation time). However this serious disruption saw the drones split around the target which subsequently resulted in the swarm being split into two smaller swarms as the drones had to move outside of their visual range to avoid the obstacle. The two swarms then continued onto their target eventually rejoining towards the end however noting that this rejoining process saw some anomaly's in the motion indicating some smoothing and refining in the swarm control could improve apparent stability. Furthermore this aligns with the large jumps in tracking errors in the x and y domains given in \autoref{fig:Linear_x_tracking_error}, \autoref{fig:Linear_y_tracking_error} and \autoref{fig:Linear_z_tracking_error} explaining its motion and confirming the robustness of the controller. \\

Furthermore observing the group metric in \autoref{fig:Linear_Group_Metric} we can observe the drones bunched up reaching their minimal safe separation as they bumped off the obstacle interference zone, however the proportional response from the controller from this near miss saw the drones separate rapidly assisting in the swarm split and further explaining the large anomaly's seen the location analysis and tracking errors. Given these factors the overall average tracking error lines plotted in \autoref{fig:Linear_x_tracking_error}, \autoref{fig:Linear_y_tracking_error} and \autoref{fig:Linear_z_tracking_error} should not be taken at face value as they are skewed by deliberate obstacle avoidance measures located in extremely challenging locations, rather the line should be used as a general metric between trials for comparison. Finally the order metric in \autoref{fig:Linear_Order_Metric} showed initially the swarm was extremely stable and was aligned with perfectly matched angular velocities, however once disrupted by the obstacle the average difference rapidly increased as expected before dropping to to a lower but slightly turbulent period as the swarm maneuvered around the obstacle and rejoined together. This metric aligns and confirms the finds from the other metrics and plots however also identifies that given a unchanging environment the swarm is very stable.
\vfill

\clearpage
\section{Figure Eight Target Performance Example}
\label{Figure Eight Target Performance Analysis}

\begin{figure} [H]
\includegraphics[width=3.2in]{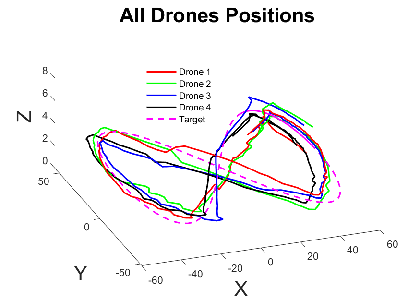}
\caption{Linear Trial 3D Plot}
\label{fig:fig8_3D Plot}
\end{figure}

\begin{figure} [H]
\includegraphics[width=3.2in]{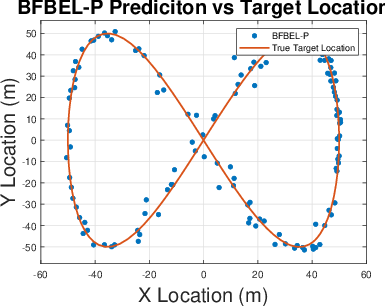}
\caption{Linear Trial BFBEL-P Performance Analysis}
\label{fig:fig8_BFBELP_Analysis}
\end{figure}

\begin{figure} [H]
\includegraphics[width=3.2in]{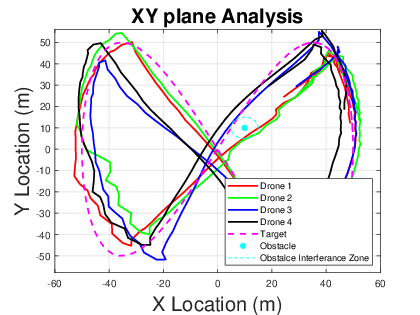}
\caption{Linear Trial xy plane with obstacle analysis}
\label{fig:fig8_xy_analysis}
\end{figure}

\begin{figure} [H]
\includegraphics[width=3.2in]{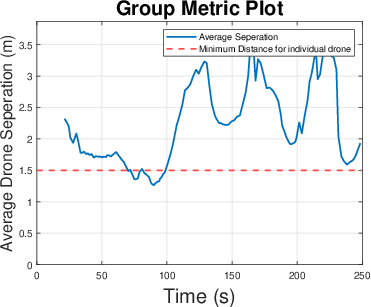}
\caption{Linear Trial Group Metric Plot}
\label{fig:fig8_Group_Metric}
\end{figure}

\begin{figure} [H]
\includegraphics[width=3.2in]{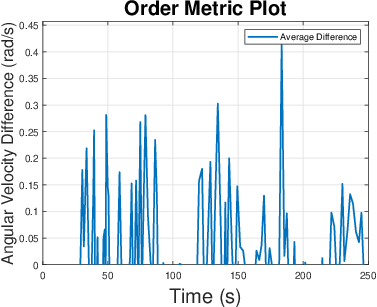}
\caption{Linear Trial Order Metric Plot}
\label{fig:fig8_Order_Metric}
\end{figure}

\begin{figure} [H]
\includegraphics[width=3.2in]{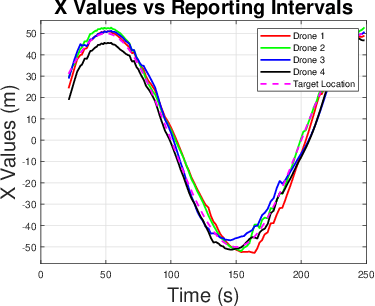}
\caption{Linear Trial x location analysis}
\label{fig:fig8_x_loc}
\end{figure}

\begin{figure} [H]
\includegraphics[width=3.2in]{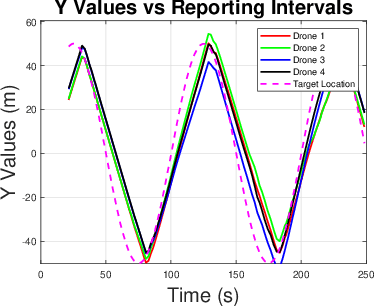}
\caption{Linear Trial y location analysis}
\label{fig:fig8_y_loc}
\end{figure}

\begin{figure} [H]
\includegraphics[width=3.2in]{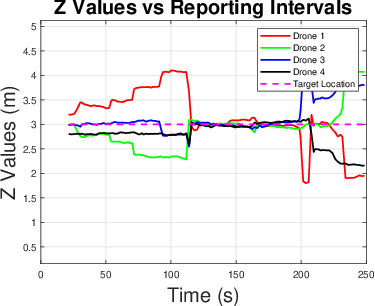}
\caption{Linear Trial z location analysis}
\label{fig:fig8_z_loc}
\end{figure}

\begin{figure} [H]
\includegraphics[width=3.2in]{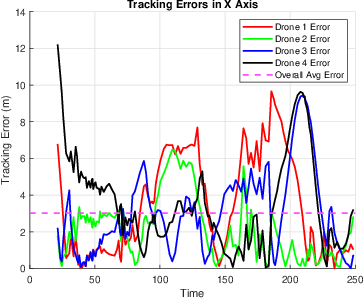}
\caption{Linear Trial x location tracking error}
\label{fig:fig8_x_tracking_error}
\end{figure}

\begin{figure} [H]
\includegraphics[width=3.2in]{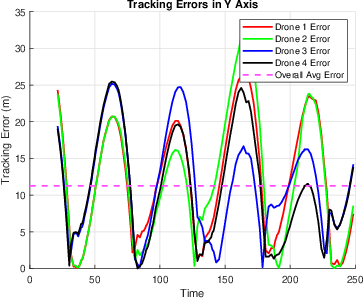}
\caption{Linear Trial y location tracking error}
\label{fig:fig8_y_tracking_error}
\end{figure}

\begin{figure} [H]
\includegraphics[width=3.2in]{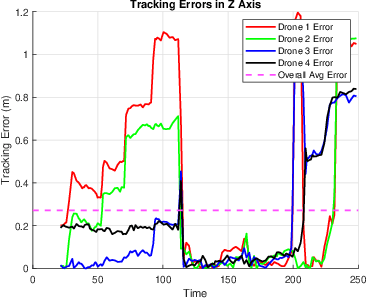}
\caption{Linear Trial z location tracking error}
\label{fig:fig8_z_tracking_error}
\end{figure}

For each Figure Eight trial the following plots are generated that being a 3D plot \autoref{fig:fig8_3D Plot}, BFBEL-P vs Target Prediction Comparison \autoref{fig:fig8_BFBELP_Analysis}, X Location Comparison \autoref{fig:fig8_x_loc}, Y Location Comparison \autoref{fig:fig8_y_loc}, Z Location Comparison \autoref{fig:fig8_z_loc}, XY Plane plot \autoref{fig:fig8_xy_analysis}, Group Metric \autoref{fig:fig8_Group_Metric}, Order Metric \autoref{fig:fig8_Order_Metric}, X Tracking Error \autoref{fig:fig8_x_tracking_error}, Y Tracking Error \autoref{fig:fig8_y_tracking_error} and Z Tracking Error \autoref{fig:fig8_z_tracking_error}. Additionally for each trial confidence intervals are generated for the BFBEL-P predictions in both the x and y axis which also gets analysed in a T test.  Additionally the average tracking errors are calculated along with the average standard deviation and variance for the drones in the x,y,z location compared to the target, this statistical analysis is given in \autoref{tab:linear_trial_stats}.

\begin{table}[H]
\begin{center}
\caption{Statistical Analysis from a linear trial}
\label{tab:linear_trial_stats_table}
\begin{tabular}{| p{3.8cm} | p{3cm} |}
\hline
Statistical Test & Result \\
\hline
BFBEL-P vs Target Confidence interval (95\%) in X Domain & [-0.5679, 0.1620]\\ 
\hline
BFBEL-P vs Target Confidence interval (95\%) in Y Domain & [-0.4918, 0.5676]\\ 
\hline
BFBEL-P vs Target T-Test in X Domain & p = 0.2732for $\alpha$ = 0.05\\ 
\hline
BFBEL-P vs Target T-Test in Y Domain & p = 0.8876 for $\alpha$ = 0.05\\ 
\hline
Tracking Error X Domain & 2.9836\\ 
\hline
Tracking Error Y Domain & 13.777\\ 
\hline
Tracking Error Z Domain & 0.29875\\ 
\hline
Average Standard Deviation for Overall Motion & 4.0306\\ 
\hline
Average Variance for Overall Motion & 19.5779\\ 
\hline
\end{tabular}
\end{center}
\end{table}

From the statistical data we can determine that neither the x or y BFBEL-P predictions are statistically significantly different from the target values, which combined with each result predicting within a meter with 95\% confidence, as reinforced by \autoref{fig:fig8_BFBELP_Analysis}. indicates strong predictive performance by the BFBEL-P Model. Additionally it shows sound tracking in the x and z axis in \autoref{fig:fig8_x_tracking_error} and \autoref{fig:fig8_z_tracking_error} however a strong but delayed response the y domain \autoref{fig:fig8_y_tracking_error}. This is due in part to two reasons, firstly it signifies an under damped PID response showing that for more complex trajectory a more aggressive controller is required to keep up. Additionally as the divergence occurs at the switch back points (peaks and troughs of \autoref{fig:fig8_y_loc}) the target was likely moving faster than the drones were speed limited to as part of their stabilization efforts. This in tern indicates that while the swarm is very stable and strong as shown by the Group and Order metrics in \autoref{fig:fig8_Group_Metric} and \autoref{fig:fig8_Order_Metric} its stability is constraining its ability to respond quick enough for this rapid maneuvering. \\

Furthermore analysing the obstacle avoidance in \autoref{fig:fig8_xy_analysis} we can see the drones avoided the target once moved into the interference zone critically noting that the target was located directly on top of where the target was at the time. However this serious disruption saw the drones split around the target which subsequently resulted in the swarm being split into two smaller swarms as the drones had to move outside of their visual range to avoid the obstacle. The two swarms then continued onto their target eventually rejoining towards the end however noting that this rejoining process saw some anomaly's in the motion indicating some smoothing and refining in the swarm control could improve apparent stability. Furthermore this aligns with the large jumps in tracking errors in the x and y domains given in \autoref{fig:fig8_x_tracking_error}, \autoref{fig:fig8_y_tracking_error} and \autoref{fig:fig8_z_tracking_error} explaining its motion and confirming the robustness of the controller. Furthermore observing the group metric in \autoref{fig:fig8_Group_Metric} we can observe the drones bunched up crossing their minimal safe separation as they hit the first switch back, however the proportional response from the controller from this near miss saw the drones separate rapidly assisting in the swarm split to avoid a collision, further explaining the large anomaly's seen the location analysis and tracking errors. Given these factors the overall average tracking error lines plotted in \autoref{fig:fig8_x_tracking_error}, \autoref{fig:fig8_y_tracking_error} and \autoref{fig:fig8_z_tracking_error} should not be taken at face value as they are skewed by deliberate obstacle avoidance measures located in extremely challenging locations, rather the line should be used as a general metric between trials for comparison. However once disrupted by the obstacle or a near miss the group metric plot \autoref{fig:fig8_Order_Metric} showed the average angular velocity difference rapidly decrease to a stable swarm. This metric along with others indicate strong swarming capability this in this test cse is slight under damped in its response in one axis, yet critically the BFBEL-P controller maintains excellent predictive capability despite the more complex motion.


\clearpage
\onecolumn
\section{RQT Plot}
\label{RQT Plot Information}
\begin{figure}[H]
\begin{centering}
    \includegraphics[width=7.5in]{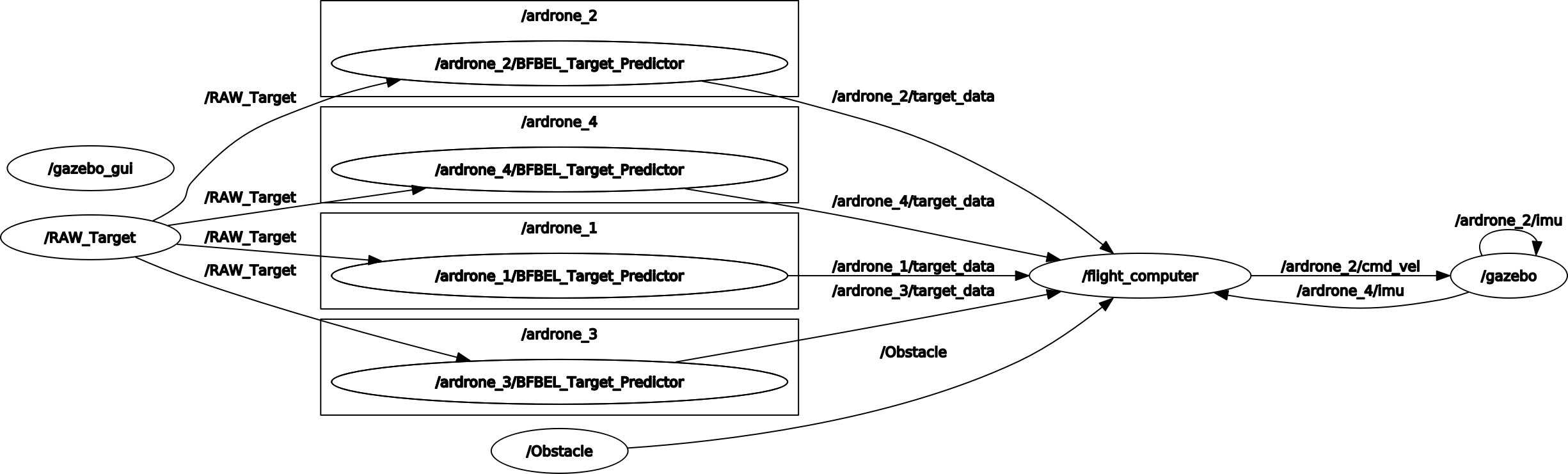}
    \caption{RQT Plot generated by the ROS simulation}
    \label{fig:rqt_plot}
\end{centering}
\end{figure}

\begin{figure}[H]
\begin{centering}
    \includegraphics[width=5in]{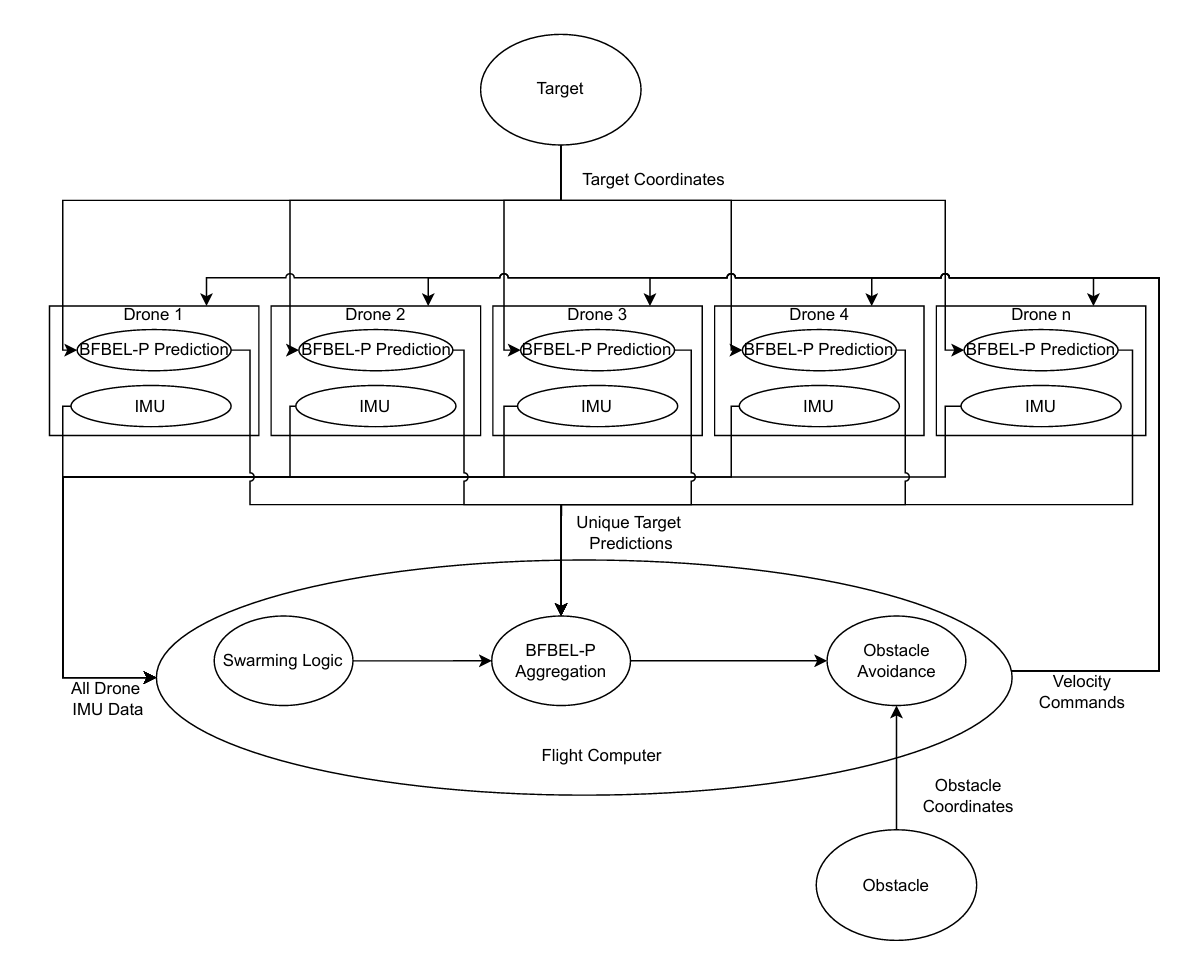}
    \caption{Custom RQT Plot to simplify data pathways}
    \label{fig:rqt_custom}
\end{centering}
\end{figure}

\begin{figure}[H]
\includegraphics[width=7in]{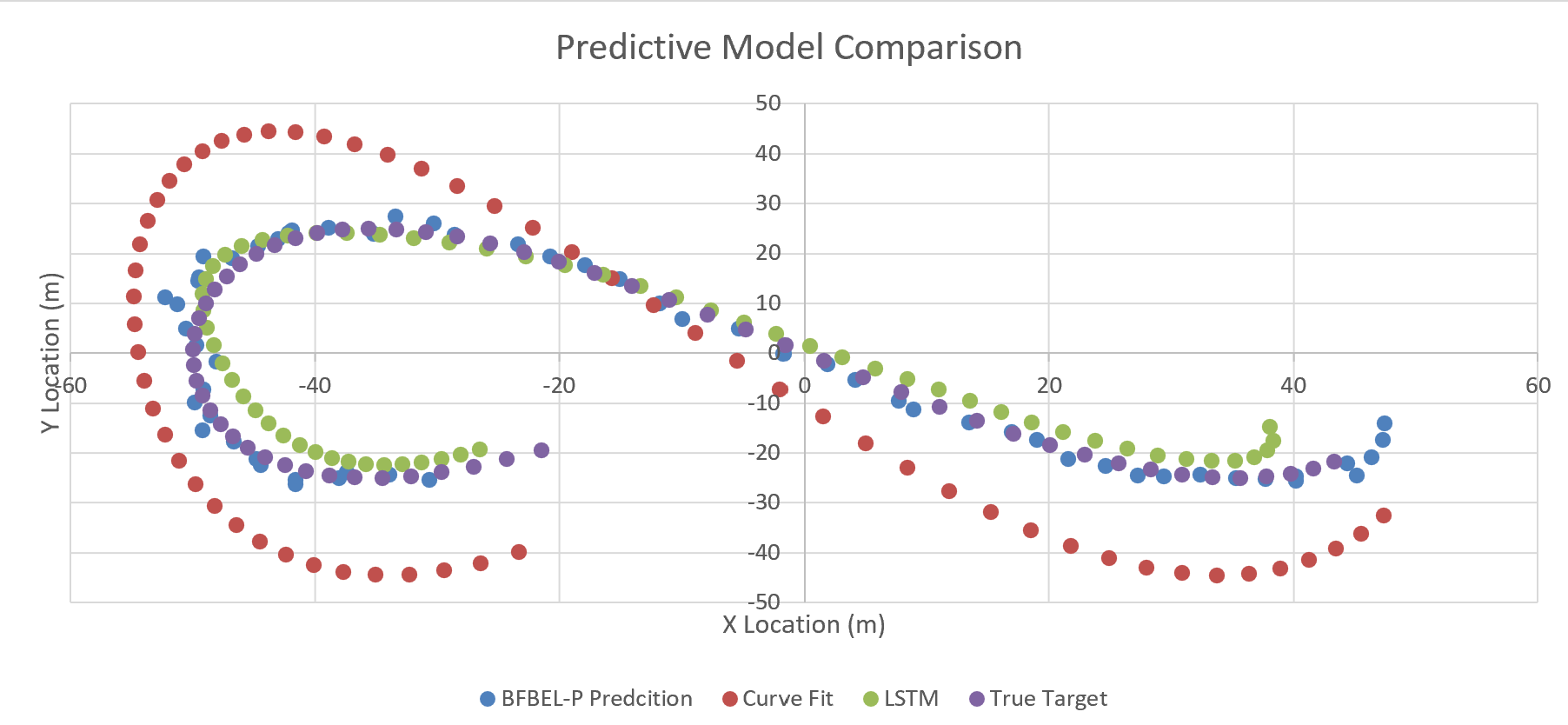}
\caption{BFBEL-P, Curve Fitting and LSTM Short Term Predictive Comparison }
\label{fig:BFBEL_Pred_comparison}
\end{figure}

\begin{figure}[H]
\includegraphics[width=7in]{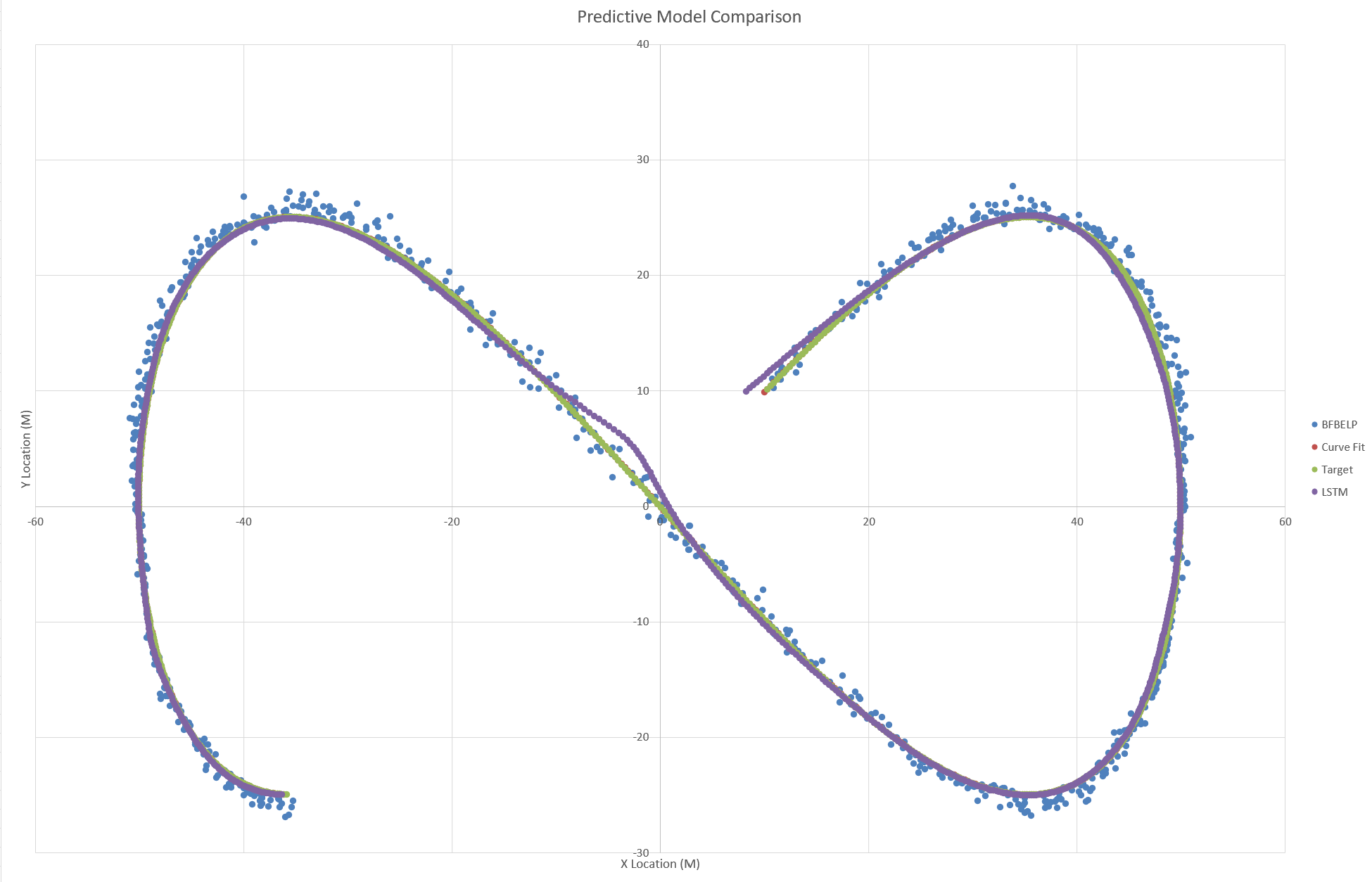}
\caption{BFBEL-P, Curve Fitting and LSTM  Long Term Predictive Comparison (Curve fit is mostly obscured underneath Target)}
\label{fig:long_term_comparison}
\end{figure}

This RQT Plot in \autoref{fig:rqt_plot} demonstrates the data pathways within the ROS simulation, critically each node within the simulation is represented by a named oval, and the data pathways are represented via arrows with the respective information (topics) attached. However noting within the simulation the drone velocity command publishing and IMU subscriptions go via the gazebo simulation to the drones due to the nature in how the Ardrone package interfaces with the drones, additionally the generated plot will only show the last drone the flight computer interacted with rather than all 4 due to the flight computer iterating though all and the RQT is taken at a snapshot in time. This in principle is identical to how the target predictions get read by the flight computer, however as the flight computer is a custom script ROS will generate the visualization the data pathway. But as the gazebo simulation is an environment it will only generate the data pathways active at that snapshot in time.
This is simplified though the  custom RQT plot \autoref{fig:rqt_custom}, which attempts to more clearly demonstrate the data pathways and processes in all stages.
\vfill

\section{Predictive Model Comparison Graphs}

\end{document}